\documentclass{article}

\usepackage{microtype}
\usepackage{graphicx}
\usepackage{subfigure}
\usepackage{booktabs} 

\usepackage{hyperref}

\usepackage[accepted]{icml2025}

\usepackage{amsmath}
\usepackage{amssymb}
\usepackage{mathtools}
\usepackage{amsthm}

\usepackage[capitalize,noabbrev]{cleveref}

\usepackage[textsize=tiny]{todonotes}

\usepackage{caption}
\usepackage{bm}
\usepackage{xcolor}
\usepackage{float}
\usepackage{multirow}
\usepackage{longtable}
\usepackage{framed}
\usepackage{enumitem}
\usepackage{pythonhighlight}
\usepackage{appendix}
\usepackage[utf8]{inputenc}
\usepackage{bbding}
\usepackage{tabularx}
\usepackage{dsfont}

\DeclareMathAlphabet{\mathpzc}{OT1}{pzc}{m}{it}

\usepackage{mathtools}
\usepackage{multirow}
\usepackage{algorithm}
\usepackage[noend]{algpseudocode}
\usepackage{color,soul}

\makeatletter
\DeclareRobustCommand\onedot{\futurelet\@let@token\@onedot}
\def\@onedot{\ifx\@let@token.\else.\null\fi\xspace}
\def\eg{\emph{e.g}\onedot} 
\def\ie{\emph{i.e}\onedot}

\def\wrt{\emph{w.r.t}\onedot} 

\makeatother

\definecolor{blue_}{RGB}{76, 114, 176}
\definecolor{orange_}{RGB}{221, 132, 82}
\definecolor{upload}{RGB}{47, 85, 151}
\definecolor{download}{RGB}{241, 13, 208}
\definecolor{red_}{RGB}{255, 0, 0}
\definecolor{gray_}{RGB}{127, 127, 127}
\definecolor{green_}{RGB}{1, 128, 0}
\definecolor{sjtured_}{RGB}{192, 0, 0}
\definecolor{sjtugreen_}{RGB}{84, 130, 53}
\definecolor{hist_red}{RGB}{194, 82, 83}
\definecolor{hist_blue}{RGB}{83, 110, 174}

\usepackage{pifont}
\usepackage{makecell}
\usepackage{colortbl}
\definecolor{grayline}{gray}{0.9}

\crefname{section}{Sec.}{Secs.}
\Crefname{section}{Section}{Sections}
\Crefname{table}{Table}{Tables}
\crefname{table}{Tab.}{Tabs.}
\crefname{figure}{Fig.}{Figs.}
\crefname{equation}{Eq.}{Eqs.}

\usepackage{wrapfig}
\usepackage{tikz}
\newcommand{\cir}[1]{\tikz[baseline]{%
    \node[anchor=base, draw, circle, inner sep=1pt, scale=0.8]{#1};}}

\usepackage{xspace}
\def\ld{\texttt{FedL2G}\xspace}
\def\ldf{\texttt{FedL2G-f}\xspace}
\def\ldl{\texttt{FedL2G-l}\xspace}

\newtheorem{assumption}{Assumption}
\newtheorem{theorem}{Theorem}
\newtheorem{lemma}{Lemma}

\icmltitlerunning{Adaptive Guidance for Local Training in Heterogeneous Federated Learning}

\begin{document}

\twocolumn[
\icmltitle{Adaptive Guidance for Local Training in Heterogeneous Federated Learning}



\icmlsetsymbol{equal}{*}

\begin{icmlauthorlist}
\icmlauthor{Jianqing Zhang}{sjtu,thu}
\icmlauthor{Yang Liu}{polyu,ailab}
\icmlauthor{Yang Hua}{qub}
\icmlauthor{Jian Cao}{sjtu,web}
\icmlauthor{Qiang Yang}{polyu}
\end{icmlauthorlist}

\icmlaffiliation{sjtu}{Shanghai Jiao Tong University}
\icmlaffiliation{polyu}{Hong Kong Polytechnic University}
\icmlaffiliation{qub}{Queen's University Belfast}
\icmlaffiliation{thu}{Institute for AI Industry Research (AIR), Tsinghua University}
\icmlaffiliation{ailab}{Shanghai Artificial Intelligence Laboratory}
\icmlaffiliation{web}{Shanghai Key Laboratory of Trusted Data Circulation and Governance in Web3}

\icmlcorrespondingauthor{Yang Liu}{yang-veronica.liu@polyu.edu.hk}
\icmlcorrespondingauthor{Jian Cao}{cao-jian@sjtu.edu.cn}

\icmlkeywords{Machine Learning, ICML}

\vskip 0.3in
]



\printAffiliationsAndNotice{}  

\begin{abstract}
Model heterogeneity poses a significant challenge in Heterogeneous Federated Learning (HtFL). In scenarios with diverse model architectures, directly aggregating model parameters is impractical, leading HtFL methods to incorporate an extra objective alongside the original local objective on each client to facilitate collaboration. However, this often results in a mismatch between the extra and local objectives. To resolve this, we propose Federated Learning-to-Guide (\ld\footnote{\url{https://github.com/TsingZ0/FedL2G}}), a method that adaptively learns to guide local training in a federated manner, ensuring the added objective aligns with each client's original goal. With theoretical guarantees, \ld utilizes only first-order derivatives \wrt model parameters, achieving a non-convex convergence rate of $\mathcal{O}(1/T)$. We conduct extensive experiments across two data heterogeneity and six model heterogeneity settings, using 14 heterogeneous model architectures (\eg, CNNs and ViTs). The results show that \ld significantly outperforms seven state-of-the-art methods.
\end{abstract}

\section{Introduction}

With the rapid development of AI techniques~\citep{touvron2023llama, achiam2023gpt}, public data has been consumed gradually, raising the need to access local data inside devices or institutions~\citep{ye2024openfedllm}. However, directly using local data often raises privacy concerns~\citep{nguyen2021federated}. Federated Learning (FL) is a promising privacy-preserving approach that enables collaborative model training across multiple clients (devices or institutions) in a distributed manner without the need to move the actual data outside clients~\citep{kairouz2019advances, li2020federated}. Nevertheless, data heterogeneity~\citep{li2021ditto, zhang2022fedala, zhang2023eliminating} and model heterogeneity~\citep{zhang2024upload, yi2023fedgh} remain two practical issues when deploying FL systems. Personalized FL (PFL) mainly focuses on the data heterogeneity issue~\citep{zhang2023pfllib}, while Heterogeneous FL (HtFL) considers both data and model heterogeneity simultaneously~\citep{zhang2024fedtgp}. HtFL's support for model heterogeneity enables a broader range of clients to participate in FL with their customized models. 

In HtFL, sharing model parameters, a widely used technique in traditional FL and PFL is not applicable~\citep{zhang2024upload}. Instead, lightweight knowledge carriers, including small auxiliary models~\citep{shen2020federated, wu2022communication, yi2024federated}, tiny homogeneous modules~\citep{liang2020think, yi2023fedgh}, and prototypes (\ie, class representative feature vectors)~\citep{jeong2018communication, tan2022fedproto}, can be shared among clients. Prototypes offer the most significant communication efficiency due to their compact size. 

\begin{figure}
	\centering
	\includegraphics[width=0.8\linewidth]{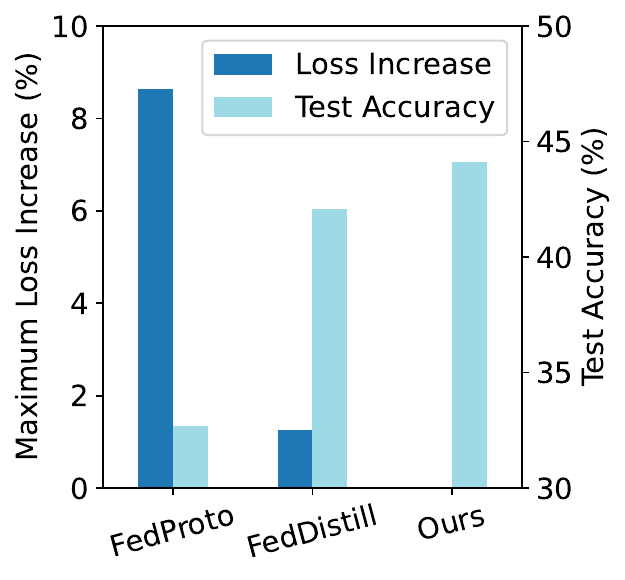}
	\caption{The objective mismatch problem increases the original local loss during FL, leading to lower test accuracy. The \textit{loss increase} is calculated as the difference between the current original local loss and its previous minimum.}
	\label{fig:intro}
\end{figure}

However, representative prototype-based methods FedDistill~\citep{jeong2018communication} and FedProto~\citep{tan2022fedproto}, still suffer from a mismatch between the prototype-guiding objective and the client's original local objective. These methods typically introduce an extra guiding objective alongside the original local objective, aiming to guide local features to align with the global ensemble prototypes. 
Due to the significant variation in width and depth among clients' heterogeneous models, their feature extraction capabilities also differ considerably~\citep{zhang2024fedtgp, zhang2024upload}. On the other hand, the data distribution also diverges across clients~\citep{mcmahan2017communication, li2022federated}. Since the global prototypes are derived from aggregating diverse local prototypes, they inherently cannot fully align with specific client models and their respective data. 
Consequently, directly optimizing the guiding and local objectives together \textit{without} prioritizing the original local objective has the potential to undermine the local objective of each client due to the objective mismatch, as shown in \cref{fig:intro}.

To address the issue of objective mismatch, we propose a novel \textbf{Federated Learning-to-Guide (\ld)} method. It prioritizes the original local objective while learning the guiding objective, ensuring that the guiding objective facilitates each client's original local task rather than causing negative effects to the original local objective. This is why we term it ``\textit{learning to guide}''. 
Specifically, we hold out a tiny \textit{quiz set} from the training set and denote the remaining set as a \textit{study set} on each client. Then we learn guiding vectors in a federated manner, ensuring that updating client models with the extra guiding loss and the original local loss on their study sets consistently reduces the original local loss on their quiz sets (which are not used for training and testing). 
The steadily decreasing original local loss (no loss increase) and the superior test accuracy illustrated in \cref{fig:intro} embody the design philosophy and effectiveness of our \ld. 
Moreover, in contrast to learning-to-learn~\citep{finn2017model, jiang2019improving, NEURIPS2020_24389bfe}, the learning-to-guide process in our \ld only requires first-order derivatives \wrt model parameters, making it computationally efficient. 

We assess the performance of our \ld across various scenarios. In addition to test accuracy, we also evaluate communication and computation overhead. The results consistently demonstrate that \ld outperforms seven state-of-the-art methods. 
We list our contributions below:

\begin{itemize}
    \item In HtFL with data and model heterogeneity, we analyze and observe the objective mismatch issue between the extra guiding objective and the original local objective within representative prototype-based methods.
    \item We propose a \ld method that prioritizes the original local objective while using the extra guiding objective to eliminate the objective mismatch issue.
    \item We prove that \ld achieves efficiency using only first-order derivatives \wrt model parameters, with a non-convex convergence rate of $\mathcal{O}(1/T)$.
    \item To demonstrate our \ld’s priority, we conducted extensive experiments covering two types of data heterogeneity, six types of model heterogeneity (including 14 distinct model architectures such as CNNs and ViTs), and various system settings. 
\end{itemize}

\section{Related Work}

\subsection{Heterogeneous Federated Learning (HtFL)}

Presently, FL is one of the popular collaborative learning and privacy-preserving techniques~\citep{zhang2022fedala, li2020federated} and HtFL extends traditional FL by supporting model heterogeneity~\citep{ye2023heterogeneous}. 
Prevailing HtFL methods primarily consider three types of model heterogeneity: (1) group heterogeneity, (2) partial heterogeneity, and (3) full heterogeneity~\citep{zhang2024upload}. 
Among them, the HtFL methods considering group model heterogeneity extract different but architecture-constraint sub-models from a global model for various groups of clients~\citep{diao2020heterofl, horvath2021fjord, wen2022federated, luo2023dfrd, zhou2024every}. Thus, they cannot support customized client models and are \textit{excluded} from our consideration. Additionally, sharing and revealing model architectures within each group of clients also raises privacy and intellectual property concerns~\citep{zhang2024fedtgp}. 
As the server is mainly utilized for parameter aggregation in prior FL systems~\citep{tan2022towards, kairouz2019advances}, training a server module with a large number of epochs, like~\citep{zhang2024upload, zhang2024fedtgp, zhu2021data}, necessitates additional upgrades or the purchase of a new heavy server, which is impractical. Thus, we focus on the \textit{server-lightweight} methods. 

Both partial and full model heterogeneity accommodate customized client model architectures, but partial heterogeneity still assumes that some small parts of all client models are homogeneous. 
For example, LG-FedAvg~\citep{liang2020think} and FedGH~\citep{yi2023fedgh} stand out as two representative approaches. LG-FedAvg and FedGH partition each client model into a feature extractor part and a classifier head part, operating under the assumption that all classifier heads are homogeneous. 
In LG-FedAvg, the parameters of classifier heads are uploaded to the server for aggregation. In contrast, FedGH uploads prototypes to the server and trains the lightweight global classifier head for a small number of epochs. 
Both methods utilize the global head for knowledge transfer among clients but overlook the inconsistency between the global head and local tasks. 

In the case of full model heterogeneity, mutual distillation~\citep{zhang2018deep} and prototype guidance~\citep{tan2022fedproto} emerge as two prevalent techniques. Using mutual distillation, FML~\citep{shen2020federated}, FedKD~\citep{wu2022communication}, and FedMRL~\citep{yi2024federated} facilitate client knowledge transfer through a globally shared auxiliary model. However, sharing an entire model demands substantial communication resources, even if the auxiliary model is typically small~\citep{zhang2024upload}. Furthermore, aggregating a global model in scenarios with data heterogeneity presents numerous challenges, such as client-drift~\citep{karimireddy2020scaffold}, ultimately leading to a subpar global model~\citep{li2022federated, zhang2023eliminating, zhang2023gpfl, Zhang2023fedcp}. 
As representative prototype guidance methods, FedDistill~\citep{jeong2018communication} and FedProto~\citep{tan2022fedproto} gather prototypes on each client, aggregate them on the server to create the global prototypes, and guide client local training with these global prototypes. Specifically, FedDistill extracts lower-dimensional prototypes than FedProto. This difference stems from FedDistill applying prototype guidance in the logit space, whereas FedProto uses the intermediate feature space. Sharing higher-dimensional prototypes can transfer more information among clients but may also exacerbate the negative effects of objective mismatch. 

\subsection{Student-Centered Guidance}

Our learning-to-guide philosophy draws inspiration from student-centered knowledge distillation approaches~\citep{yang2024learning}. They are based on the insight that a teacher's subject matter expertise alone may not match the student's specific studying ability and style, resulting in negative effects~\citep{sengupta2023good, yang2024learning}. To address the mismatch between the teacher's knowledge and the needs of the student, updating the teacher model with concise feedback from the student on a small quiz set represents a promising direction~\citep{ma2022knowledge, zhou2022bert, sengupta2023good}. 

However, these student-centered approaches are built upon a teacher-student framework, assuming the presence of a well-trained large teacher model. They concentrate on a central training scheme without factoring in distributed multiple students and privacy protection~\citep{lee2022meta, hu2022teacher}, rendering them inapplicable in the context of HtFL. Additionally, modifying and extending these student-centered approaches to HtFL requires significant communication and computational resources to update a shared large teacher model based on student feedback~\citep{zhou2022bert, lu2023meta}.
Nevertheless, the student-centered guidance concept inspires us to propose a learning-to-guide approach in HtFL. This involves substituting the large teacher model with compact guiding vectors and updating these guiding vectors based on clients’ feedback from their quiz sets, making our \ld lightweight, efficient, and adaptable. 

\section{Federated Learning-to-Guide: \ld}

\subsection{Notations and Preliminaries}

\noindent\textbf{Problem statement. \ } In an HtFL system, $N$ clients, on the one hand, train their heterogeneous local models (with parameters ${\bm \theta}_1, \ldots, {\bm \theta}_N$) using their private and heterogeneous training data $\mathcal{D}_1, \ldots, \mathcal{D}_N$. On the other hand, they share some global information, denoted by $\mathcal{G}$, with the assistance of a server to facilitate collaborative learning. Formally, the typical objective of HtFL is
\begin{equation}
    \min_{{\bm \theta}_1, \ldots, {\bm \theta}_N} \ \sum_{i=1}^N \frac{|\mathcal{D}_i|}{D} \mathcal{L}_{\mathcal{D}_i}({\bm \theta}_i, \mathcal{G}),
\end{equation}
where $|\mathcal{D}_i|$ represents the size of the training set $\mathcal{D}_i$, $D=\sum_{i=1}^N |\mathcal{D}_i|$, and $\mathcal{L}_{\mathcal{D}_i}$ denotes a total client training objective over $\mathcal{D}_i$. 

\noindent\textbf{Prototype-based HtFL. \ } Sharing class-wise prototypes of low-dimensional features in either the intermediate feature space or the logit space among clients has become a prevalent and communication-efficient solution to address model heterogeneity in HtFL~\citep{ye2023heterogeneous}. Take the popular scheme~\citep{jeong2018communication} for example, where prototypes are shared in the logit space, $\mathcal{G}$ (the set of global prototypes) is defined by
\begin{equation}
\begin{aligned}
    \mathcal{G} = \{{\bm g}^y\}_{y=1}^C, \quad {\bm g}^y = agg(\{{\bm g}^y_1, \ldots, {\bm g}^y_N\}), 
\end{aligned}
\end{equation}
where ${\bm g}^y_i = \mathbb{E}_{\mathcal{D}_{i, y}} [f_i({\bm x}, {\bm \theta}_i)]$, $\mathbb{E}_{\mathcal{D}}$ is short for $\mathbb{E}_{({\bm x}, y)\sim \mathcal{D}}$ for any $\mathcal{D}$ and $C$ represents the total number of clients' original local task classes. ${\bm g}^y$ and ${\bm g}^y_i$ denote the global and local prototypes of class $y$, respectively. Besides, $agg$ is an aggregation function defined by each prototype-based HtFL method, $\mathcal{D}_{i, y}$ stands for a subset of $\mathcal{D}_i$ containing all the data of class $y$, and $f_i$ represents the local model of client $i$. Given a global $\mathcal{G}$, client $i$ then takes prototype guidance for knowledge transfer among clients via
\begin{equation}
    \mathcal{L}_{\mathcal{D}_i}({\bm \theta}_i, \mathcal{G}) := \mathbb{E}_{\mathcal{D}_i} [\ell_{ce}(f_i({\bm x}, {\bm \theta}_i), y) + \ell_{g}(f_i({\bm x}, {\bm \theta}_i), {\bm g}^y)], \label{eq:kd}
\end{equation}
where the weight of $\ell_{g}$ is set to one to balance two objectives equally here, $\ell_{ce}$ is the original local cross-entropy loss~\citep{zhang2018generalized}, and $\ell_{g}$ is the guiding loss. 

\subsection{Learning to Guide}
\label{sec:LTG}

\noindent\textbf{Motivation. \ }
Initially, heterogeneous client models trained by $\ell_{ce}$ can adapt to their local data with diverse feature extraction capabilities. However, directly adding $\ell_{g}$ \textit{without} prioritizing $\ell_{ce}$ can cause the model of each client to deviate from $\ell_{ce}$. 
On the other hand, since all feature vectors are extracted on heterogeneous client data, the aggregated global prototype, \eg, ${\bm g}^y$, is data-derived, which may deviate from the features regarding class $y$ on each client. 
Both the model and data heterogeneity result in the objective mismatch issue between $\ell_{ce}$ and $\ell_{g}$, which causes the negative effect to $\ell_{ce}$ when using $\ell_{g}$, as shown in \cref{fig:intro} and discussed further in \cref{sec:property}. 
Therefore, we propose a novel \ld method, which substitutes the data-derived prototypes with trainable guiding vectors $\mathcal{G} = \{{\bm v}^y\}_{y=1}^C$ and ensures that \textbf{\textit{$\mathcal{G}$ is learned to reduce $\ell_{ce}$ when guided by $\ell_{g}$}}. Formally, we replace \cref{eq:kd} with a new loss to train the client model:
\begin{equation}
    \mathcal{L}_{\mathcal{D}_i}({\bm \theta}_i, \mathcal{G}) := \mathbb{E}_{\mathcal{D}_i} [\ell_{ce}(f_i({\bm x}, {\bm \theta}_i), y) + \ell_{g}(f_i({\bm x}, {\bm \theta}_i), {\bm v}^y)],  \label{eq:kd_new}
\end{equation}
where the learning of guiding vectors $\mathcal{G}$ is the key step. 

\noindent\textbf{Learning guiding vectors. \ } Without relying on data-derived information, we randomly initialize the global $\mathcal{G}$ on the server and update it based on the aggregated gradients from participating clients in each communication iteration. Inspired by the technique of outer-inner loops in meta-learning~\citep{zhou2022bert}, we derive the gradients of client-specific ${\bm v}^y_i$ in the \textit{outer-loop}, while focusing on reducing the original local loss, \ie, $\ell_{ce}$, in the \textit{inner-loop} on each client. 
To implement the learning-to-guide process, we hold out a tiny \textit{quiz set} $\mathcal{D}^q_i$ (one batch of data) from $\mathcal{D}_i$ and denote the remaining training set as the \textit{study set} $\mathcal{D}^s_i$. Notice that we exclusively conduct model updates on $\mathcal{D}^s_i$ and never train ${\bm \theta}_i$ on $\mathcal{D}^q_i$. In particular, $\mathcal{D}^q_i$ is solely used to evaluate ${\bm \theta}_i$'s performance regarding the original local loss and derive the gradients (feedback) \wrt ${\bm v}^y_i$. Below, we describe the details of \ld in the $t$-th iteration, using the notation $t$ solely for the global $\mathcal{G}$ for clarity. 
Recall that $\mathcal{G} = \{{\bm v}^y\}_{y=1}^C$, we use the general notation $\mathcal{G}$ in the following descriptions for simplicity, although all operations correspond to each ${\bm v}^y, y \in \{1, \ldots, C\}$ within $\mathcal{G}$. 

Firstly, in step \textbf{\cir{1}}, we download $\mathcal{G}^{t-1}$ from the server to client $i$. Then, in step \textbf{\cir{2}}, we perform regular training for ${\bm \theta}_i$ on $\mathcal{D}^s_i$ using $\mathcal{L}_{\mathcal{D}^s_i}({\bm \theta}_i, \mathcal{G}^{t-1})$ (see \cref{eq:kd_new}). Sequentially, the pivotal steps \textbf{\cir{3}} and \textbf{\cir{4}} correspond to our objective of learning-to-guide. 
In step \textbf{\cir{3}}, we execute a \textit{pseudo-train} step (without saving the updated model back to disk) on a randomly sampled batch $\mathcal{B}^s_i$ from $\mathcal{D}^s_i$, \ie, 
\begin{equation}
    {\bm \theta}'_i(\mathcal{G}^{t-1}) \leftarrow {\bm \theta}_i - \eta_c \nabla_{{\bm \theta}_i} \mathcal{L}_{\mathcal{B}^s_i}({\bm \theta}_i, \mathcal{G}^{t-1}), \label{eq:pseudo-train}
\end{equation}
where $\eta_c$ is the client learning rate, and we call ${\bm \theta}'_i(\mathcal{G}^{t-1})$ as the pseudo-trained local model parameters, which is a function of $\mathcal{G}^{t-1}$. In step \textbf{\cir{4}}, our aim is to update the $\mathcal{G}^{t-1}$ in $\mathcal{L}_{\mathcal{B}^s_i}({\bm \theta}_i, \mathcal{G}^{t-1})$ (see \cref{eq:kd_new}) to minimize $\ell_{ce}$ with ${\bm \theta}'_i(\mathcal{G}^{t-1})$ on $\mathcal{D}^q_i$, thus we compute the gradients of $\mathcal{G}^{t-1}$ \wrt $\ell_{ce}$ on $\mathcal{D}^q_i$: $\nabla_{\mathcal{G}^{t-1}} \mathbb{E}_{\mathcal{D}^q_i} [\ell_{ce}(f_i({\bm x}, {\bm \theta}'_i(\mathcal{G}^{t-1})), y)]$ (see \cref{sec:eff} for details). Afterwards, we upload clients' gradients of $\mathcal{G}^{t-1}$ in step \textbf{\cir{5}} and aggregate them in step \textbf{\cir{6}}. Then, in step \textbf{\cir{7}}, we update the global $\mathcal{G}^{t-1}$ on the server with the aggregated gradients. Put steps \textbf{\cir{3}}, \textbf{\cir{4}}, \textbf{\cir{5}}, \textbf{\cir{6}}, \textbf{\cir{7}} together, we have
\begin{equation}
\begin{aligned}
    \mathcal{G}^t = \mathcal{G}^{t-1} - \eta_s \frac{1}{|\mathcal{I}^t|} &\sum_{i\in \mathcal{I}^t} \nabla_{\mathcal{G}^{t-1}} \mathbb{E}_{\mathcal{D}^q_i} [\ell_{ce}(f_i({\bm x}, {\bm \theta}_i \\&- \eta_c \nabla_{{\bm \theta}_i} \mathcal{L}_{\mathcal{B}^s_i}({\bm \theta}_i, \mathcal{G}^{t-1})), y)], \label{eq:agg_grad}
\end{aligned}
\end{equation}
where $\eta_s$ is the server learning rate and $\mathcal{I}^t$ is the set of participating clients in the $t$-th iteration. We utilize the weight $\frac{1}{|\mathcal{I}^t|}$ here, considering that all participating clients execute step \textbf{\cir{3}} and \textbf{\cir{4}} with identical sizes of $\mathcal{B}^s_i$ and $\mathcal{D}^q_i$, $i \in \{1, \ldots, N\}$. Since some classes may be absent on certain clients, we only upload and aggregate the non-zero gradient vectors to minimize communication costs. We can easily implement \cref{eq:agg_grad} using popular public tools, \eg, higher~\citep{grefenstette2019generalized}. 


\noindent\textbf{Warm-up period. \ } Since $\mathcal{G}$ is randomly initialized, using an uninformative $\mathcal{G}$ misguides local model training in \cref{eq:kd_new}. Thus, before conducting regular client training in step \textbf{\cir{2}}, \ld requires a warm-up period of $T'$ (see more analysis of $T'$ in \cref{sec:hyper}, where \ld also performs well \textit{without} warming-up) iterations with step \textbf{\cir{1}}, \textbf{\cir{3}}, \textbf{\cir{4}}, \textbf{\cir{5}}, \textbf{\cir{6}}, \textbf{\cir{7}}. Without step \textbf{\cir{2}}, the warm-up process only involves one batch of each client's quiz set, thus demanding relatively small computation overhead. 

\noindent\textbf{Twin HtFL methods based on \ld. \ } The above processes assume sharing information in the logit space, denoted as \ldl. Additionally, when considering the intermediate feature space, we can rephrase all the corresponding $\ell_{g}$, for instance, rewriting $\ell_{g}(h_i({\bm x}, {\bm \theta}^h_i), {\bm v}^y)$ in \cref{eq:kd_new}, where $h_i$ represents the feature extractor component in $f_i$, ${\bm \theta}^h_i \subset {\bm \theta}_i$ denotes the associated model parameters, and ${\bm v}^y$ resides in the intermediate feature space. We denote this twin method as \ldf. The server learning rate $\eta_s$ is the \textit{unique} hyperparameter in our \ldl or \ldf. 
Due to space constraints, we offer a detailed algorithm in \cref{algo}.

\subsection{Efficiency Analysis}
\label{sec:eff}

As we compute gradients for two different entities in the \textit{outer-loop} and \textit{inner-loop}, respectively, we eliminate the necessity for calculating the second-order gradients of model parameters \wrt $\ell_{ce}$ as well as the associated computationally intensive Hessian~\citep{fallah2020convergence}. Our analysis is founded on \cref{ass:1} and \cref{ass:2} in \cref{sec:theo}. Due to space limit, we leave the derivative details to \cref{eq:4} and show client $i$'s gradient \wrt $\mathcal{G}$ here: 
\begin{equation}
    \pi_i = - \eta_c \mathbb{E}_{\mathcal{D}^q_i} \{\nabla_1 \ell_{ce} \cdot \nabla_2 f_i \cdot \mathbb{E}_{\mathcal{B}^s_i} [\nabla_2 f_i \cdot \textcolor{red_}{\nabla_{\mathcal{G}^{t-1}} \nabla_1 \ell_{g}}]\}, \label{eq:grad}
\end{equation}
where $\nabla_1 \ell_{ce} := \nabla_{a_1} \ell_{ce}(a_1, a_2)$, indicating the derivative of $\ell_{ce}$ \wrt the first variable, and so for $\nabla_2 f_i$ and $\nabla_1 \ell_{g}$. The operation $\cdot$ denotes multiplication. Computing $\nabla_1 \ell_{ce}$ and $\nabla_2 f_i$ is a common practice in deep learning~\citep{zhang2018generalized} and calculating the $\textcolor{red_}{\nabla_{\mathcal{G}^{t-1}} \nabla_1 \ell_{g}}$ term is pivotal. In alignment with existing prototype-based methods~\cite{tan2022fedproto}, the $\ell_{g}$ refers to the MSE loss, so $\ell_{g}(f_i({\bm x}', {\bm \theta}_i), {\bm v}^{y'}) = \frac{1}{M} \sum_{m=1}^M [f_i({\bm x}', {\bm \theta}_i)_m - {\bm v}^{y'}_m]^2$, where $M$ is the dimension of ${\bm v}^{y'}$. Given $\mathcal{G} = \{{\bm v}^y\}_{y=1}^C$, we have
\begin{equation}
    \textcolor{red_}{\nabla_{\mathcal{G}^{t-1}} \nabla_1 \ell_{g}} = \frac{2}{M} \sum_{m=1}^M \nabla_{\mathcal{G}^{t-1}}(f_i({\bm x}', {\bm \theta}_i)_m - {\bm v}^{y'}_m) = \textcolor{red_}{-2}. 
\end{equation}
Finally, we obtain 
\begin{equation}
    \pi_i = 2 \eta_c \mathbb{E}_{\mathcal{D}^q_i} \{\nabla_1 \ell_{ce} \cdot \nabla_2 f_i \cdot \mathbb{E}_{({\bm x}', y')\sim \mathcal{B}^s_i} [\nabla_2 f_i]\}, 
\end{equation}
where only first-order derivatives of $f_i$ \wrt ${\bm \theta}_i$ are required. 

\subsection{Convergence Analysis}

The convergence analysis of HtFL typically considers an arbitrary client, incorporating global information (\eg, $\mathcal{G}$) to facilitate collaboration~\cite{tan2022fedproto, yi2024federated}. 
Given standard assumptions in \cref{sec:theo}, we have

\begin{theorem}[One-iteration deviation]
    Let \cref{ass:1} to \cref{ass:3} hold. For an arbitrary client, after every communication iteration (with $\mathcal{G}$ for collaboration), we have
    $$
    \begin{aligned}
    \mathbb{E}[\mathcal{L}^{(t+1)E+1/2}] \le \mathcal{L}^{tE+1/2} + \frac{L_1\eta_c^2}{2} \sum_{e=1/2}^{E-1} ||\nabla \mathcal{L}^{tE+e}||_2^2 \\- \eta_c \sum_{e=1/2}^{E-1} ||\nabla \mathcal{L}^{tE+e}||_2^2 + \frac{L_1E\eta_c^2\sigma^2}{2} + 2\eta_c^2\eta_sR'ER.
    \end{aligned}
    $$ \label{th:1}
\end{theorem}

\begin{theorem}[Non-convex convergence rate of \ld]
    Let \cref{ass:1} to \cref{ass:3} hold and $\Delta = \mathcal{L}^0 - \mathcal{L}^*$, where $\mathcal{L}^*$ refers to the local optimum. Given \cref{th:1}, for an arbitrary client and an arbitrary constant $\epsilon$, our \ld has a non-convex convergence rate $\mathcal{O}(1/T)$ with 
    $$
    \begin{aligned}
    \frac{1}{T} \sum_{t=0}^{T-1} \sum_{e=1/2}^{E-1} &\mathbb{E}[||\nabla \mathcal{L}^{tE+e}||_2^2] \le \\ & \frac{\frac{2\Delta}{T} + L_1E\eta_c^2\sigma^2 + 4\eta_c^2\eta_sR'ER}{2\eta_c - L_1\eta_c^2} < \epsilon,
    \end{aligned}
    $$
    where $0 < \eta_c < \frac{2\epsilon}{L_1(E\sigma^2 + \epsilon) + 4\eta_sR'ER}$ and $\eta_s > 0$. \label{th}
\end{theorem}


\section{Experiments}

\subsection{Setup}

To evaluate the performance of our \ldl and \ldf alongside 7 popular \textit{server-lightweight} HtFL methods: LG-FedAvg~\citep{liang2020think}, FedGH~\citep{yi2023fedgh}, FML~\citep{shen2020federated}, FedKD~\citep{wu2022communication}, FedMRL~\citep{yi2024federated}, FedDistill~\citep{jeong2018communication}, and FedProto~\citep{tan2022fedproto}, we conduct comprehensive experiments on four public datasets under two widely used data heterogeneity settings, involving up to 14 heterogeneous model architectures. 
Specifically, we demonstrate the encouraging performance of \ld in accuracy, communication cost, and computation cost. Subsequently, we investigate the characteristics behind our \ld from an experimental perspective. 


\noindent\textbf{Data heterogeneity settings. \ } 
Following existing work~\citep{zhang2022fedala, lin2020ensemble, zhang2023gpfl, zhang2024fedtgp}, we adopt two popular settings across four enduring datasets Cifar10~\citep{krizhevsky2009learning}, Cifar100~\citep{krizhevsky2009learning}, Flowers102~\citep{nilsback2008automated}, and Tiny-ImageNet~\citep{chrabaszcz2017downsampled}. 
Concretely, we simulate pathological data heterogeneity settings by allocating sub-datasets with 2/10/10/20 data classes from Cifar10/Cifar100/Flowers102/Tiny-ImageNet to each client. In Dirichlet data heterogeneity settings, we allocate the data of class $y$ to each client using a client-specific ratio $q^y$ from a given dataset. $q^y$ is sampled from a Dirichlet distribution with a control parameter $\beta$ as described in~\citep{lin2020ensemble}. By default, we set $\beta=0.1$ for Cifar10 and Cifar100, and $\beta=0.01$ for Flowers102 and Tiny-ImageNet to enhance setting diversity. In both the pathological and Dirichlet settings, the data quantity among clients varies to account for unbalanced scenarios. 

\noindent\textbf{Model heterogeneity settings. \ } 
To neatly denote model heterogeneity settings, we utilize the notation HtFE$_X$ following the convention in~\citep{zhang2024upload} to represent a group of heterogeneous feature extractors, where $X$ denotes the degree of model heterogeneity (positive correlation), while the remaining classifier heads remain homogeneous. For example, HtFE$_8$ denotes a group of eight heterogeneous feature extractors from eight model architectures (4-layer CNN~\citep{mcmahan2017communication}, GoogleNet~\citep{szegedy2015going}, MobileNet\_v2~\citep{sandler2018mobilenetv2}, ResNet18, ResNet34, ResNet50, ResNet101, and ResNet152~\citep{he2016deep}), respectively. In addition, we use the notation HtM$_X$ to denote a group of fully heterogeneous models. Within a specific group, for instance, HtFE$_X$, we allocate the $(i \mod X)$th model in this group to client $i$ with reinitialized parameters. Given the popularity of all models within HtFE$_8$ in the FL field, our primary focus is on utilizing HtFE$_8$. Additionally, some baseline methods, such as LG-FedAvg and FedGH, assume the classifier heads to be homogeneous, making HtM$_X$ inapplicable for them. Moreover, to meet the prerequisite of identical feature dimensions ($K$) for FedGH, FedKD, and FedProto, we incorporate an average pooling layer~\citep{szegedy2015going} before the classifier heads and set $K=512$ for all models. 

\begin{table*}[h]
  \centering
  \setlength{\tabcolsep}{5pt}
    \caption{The test accuracy (\%) on four datasets in two data heterogeneity settings using HtFE$_8$.}
  \resizebox{!}{!}{
    \begin{tabular}{l|*{4}{c}|*{4}{c}}
    \toprule
    Settings & \multicolumn{4}{c|}{Pathological Setting} & \multicolumn{4}{c}{Dirichlet Setting} \\
    \midrule
    Datasets & C10 & C100 & F102 & TINY & C10 & C100 & F102 & TINY \\
    \midrule
    LG-FedAvg & 86.8$\pm$.3 & 57.0$\pm$.7 & 58.9$\pm$.3 & 32.0$\pm$.2 & 84.6$\pm$.5 & 40.7$\pm$.1 & 70.0$\pm$.9 & 48.2$\pm$.1 \\
    FedGH & 86.6$\pm$.2 & 57.2$\pm$.2 & 59.3$\pm$.3 & 32.6$\pm$.4 & 84.4$\pm$.3 & 41.0$\pm$.5 & 69.7$\pm$.2 & 46.7$\pm$.1 \\
    FML & 87.1$\pm$.2 & 55.2$\pm$.1 & 57.8$\pm$.3 & 31.4$\pm$.2 & 85.9$\pm$.1 & 39.9$\pm$.3 & 68.4$\pm$1.2 & 47.1$\pm$.1 \\
    FedKD & 87.3$\pm$.3 & 56.6$\pm$.3 & 54.8$\pm$.4 & 32.6$\pm$.4 & 86.5$\pm$.2 & 40.6$\pm$.3 & 69.6$\pm$1.6 & 48.2$\pm$.5 \\
    FedMRL & 87.8$\pm$.3 & 59.8$\pm$.5 & 60.9$\pm$.8 & 33.2$\pm$.4 & 86.2$\pm$.4 & 41.2$\pm$.5 & 70.1$\pm$.7 & 48.2$\pm$.9 \\
    FedDistill & 87.2$\pm$.1 & 57.0$\pm$.3 & 58.5$\pm$.3 & 31.5$\pm$.4 & 86.0$\pm$.3 & 41.5$\pm$.1 & 71.2$\pm$.7 & 48.8$\pm$.1 \\
    FedProto & 83.4$\pm$.2 & 53.6$\pm$.3 & 55.1$\pm$.2 & 29.3$\pm$.4 & 82.1$\pm$1.7 & 36.3$\pm$.3 & 62.3$\pm$.6 & 40.0$\pm$.1 \\
    \midrule
    \ldl & 87.7$\pm$.1 & 59.2$\pm$.4 & 60.3$\pm$.9 & 32.8$\pm$.7 & 86.5$\pm$.1 & 42.3$\pm$.1 & 71.5$\pm$.5 & 49.5$\pm$.3 \\
    \ldf & \textbf{89.3$\pm$.2} & \textbf{64.2$\pm$.3} & \textbf{64.2$\pm$.2} & \textbf{34.7$\pm$.3} & \textbf{87.6$\pm$.2} & \textbf{43.8$\pm$.4} & \textbf{73.6$\pm$.3} & \textbf{50.3$\pm$.4} \\
    \bottomrule
    \end{tabular}}
    \label{tab:data}
\end{table*}

\noindent\textbf{Other necessary settings. \ } 
Following common practice~\citep{mcmahan2017communication}, we execute one local training epoch with a batch size of 10, \ie, $\lfloor \frac{n_i}{10} \rfloor$ update steps, during each communication iteration. We conduct each experiment for up to 1000 iterations across three trials, employing a client learning rate ($\eta_c$) of 0.01, and present the best results with error bars. Moreover, we examine full participation ($\rho=1$), for 20 clients, while setting partial participation ($\rho\le 0.5$) for scenarios involving 50 and 100 clients. 
We split all client data into a training set and a test set for each client at a ratio of 75\% and 25\%, respectively, and we evaluate the averaged test accuracy on clients' test sets. 
Please refer to the \cref{sec:addexp} for more details and results. 

\subsection{Performance of \ld}
\label{sec:perform}

\subsubsection{Data Heterogeneity Settings}

To save space, we utilize brief abbreviations to represent the dataset names, specifically: ``C10'' for Cifar10, ``C100'' for Cifar100, ``F102'' for Flowers102, and ``TINY'' for Tiny-ImageNet. Based on \cref{tab:data}, both \ldl and \ldf show superior performance compared to baseline methods. Notably, \ldf demonstrates better performance across all datasets and data heterogeneity scenarios. This can be attributed to the fact that \ldl learns to guide the original local task in the logit space, while \ldf focuses on the intermediate feature space, and the latter contains richer information due to its higher dimension. 
Regarding accuracy, \ldf surpasses the best baseline FedGH on Cifar100 by \textbf{4.4\%} in the pathological setting. Methods based on mutual distillation, such as FML, FedKD, and FedMRL, transfer more information (with more bits) than other methods in each iteration. Yet, they do not consistently achieve optimal performance due to the absence of a teacher model with prior knowledge. 
FedMRL achieves better performance by combining global and local models during inference, though this results in increased inference overhead. 
FedProto suffers in the model heterogeneity setting and performs the worst, as client models exhibit varying feature extraction abilities~\citep{zhang2024fedtgp}. Conversely, our \ldf excels with learning-to-guide in the intermediate feature space. While FedDistill mitigates this issue by sharing prototypical logits, there is still room for improvement through learning-to-guide in the logit space, a capability offered by \ldl. 

\subsubsection{Various Model Heterogeneity Degrees}

\begin{table*}[h]
  \centering
    \caption{The test accuracy (\%) on Cifar100 in the default Dirichlet setting with incremental degrees of model heterogeneity or more clients. ``(a, b)'' represents (client amount $N$, client participation ratio $\rho$).}
  \resizebox{!}{!}{
    \begin{tabular}{l|*{5}{c}|*{3}{c}}
    \toprule
    Settings & \multicolumn{5}{c|}{Incremental Degrees of Model Heterogeneity} & \multicolumn{3}{c}{More Clients}\\
    \midrule
    & HtFE$_2$ & HtFE$_3$ & HtFE$_4$ & HtFE$_9$ & HtM$_{10}$ & (50, 0.5) & (100, 0.5) & (100, 0.1) \\
    \midrule
    LG-FedAvg & 46.6$\pm$.2 & 45.6$\pm$.4 & 43.9$\pm$.2 & 42.0$\pm$.3 & --- & 37.8$\pm$.1 & 35.1$\pm$.5 & 41.0$\pm$.2 \\
    FedGH & 46.7$\pm$.4 & 45.2$\pm$.2 & 43.3$\pm$.2 & 43.0$\pm$.9 & --- & 37.3$\pm$.4 & 34.3$\pm$.2 & 40.3$\pm$.8 \\
    FML & 45.9$\pm$.2 & 43.1$\pm$.1 & 43.0$\pm$.1 & 42.4$\pm$.3 & 39.9$\pm$.1 & 38.8$\pm$.1 & 36.1$\pm$.3 & 35.2$\pm$.9 \\
    FedKD & 46.3$\pm$.2 & 43.2$\pm$.5 & 43.2$\pm$.4 & 42.3$\pm$.4 & 40.4$\pm$.1 & 38.3$\pm$.4 & 35.6$\pm$.6 & 36.5$\pm$.2 \\
    FedMRL & 46.6$\pm$.4 & 44.5$\pm$.6 & 44.2$\pm$.2 & 43.9$\pm$.4 & 42.1$\pm$.1 & 38.6$\pm$.2 & 36.4$\pm$.6 & 41.7$\pm$.3 \\
    FedDistill & 46.9$\pm$.1 & 43.5$\pm$.2 & 43.6$\pm$.1 & 42.1$\pm$.2 & 41.0$\pm$.1 & 38.5$\pm$.4 & 36.1$\pm$.2 & 41.2$\pm$.5 \\
    FedProto & 44.0$\pm$.2 & 38.1$\pm$.6 & 34.7$\pm$.6 & 32.7$\pm$.8 & 36.1$\pm$.1 & 33.0$\pm$.4 & 29.0$\pm$.5 & 28.6$\pm$.9 \\
    \midrule
    \ldl & 47.3$\pm$.1 & 44.5$\pm$.1 & \textbf{44.8$\pm$.1} & 44.1$\pm$.1 & 41.8$\pm$.2 & 38.9$\pm$.2 & 36.7$\pm$.1 & 41.6$\pm$.4 \\
    \ldf & \textbf{47.8$\pm$.3} & \textbf{45.8$\pm$.1} & 44.7$\pm$.1 & \textbf{45.7$\pm$.2} & \textbf{43.5$\pm$.1} & \textbf{40.5$\pm$.0} & \textbf{37.9$\pm$.3} & \textbf{42.3$\pm$.7} \\
    \bottomrule
    \end{tabular}}
    \label{tab:model}
\end{table*}

Besides the HtFE$_8$ group, we also explore 5 other model heterogeneity settings, while maintaining consistent data heterogeneity in the Dirichlet setting to control variables. The degree of model heterogeneity escalates from HtFE$_2$ to HtM$_{10}$ as follows: HtFE$_2$ comprises 4-layer CNN and ResNet18; HtFE$_3$ includes ResNet10~\citep{zhong2017deep}, ResNet18, and ResNet34; HtFE$_4$ comprises 4-layer CNN, GoogleNet, MobileNet\_v2, and ResNet18; HtFE$_9$ includes ResNet4, ResNet6, and ResNet8~\citep{zhong2017deep}, ResNet10, ResNet18, ResNet34, ResNet50, ResNet101, and ResNet152; HtM$_{10}$ contains all the model architectures in HtFE$_8$ plus two additional architectures ViT-B/16~\citep{dosovitskiy2020image} and ViT-B/32~\citep{dosovitskiy2020image}. ViT models have a complex classifier head, whereas other CNN-based models only consider the last fully connected layer as the classifier head. Consequently, methods assuming a homogeneous classifier head, such as LG-FedAvg and FedGH, do not apply to HtM$_{10}$. Referring to \cref{tab:model}, our \ldl and \ldf still perform well in these scenarios, particularly in more model-heterogeneous settings. As the setting becomes more heterogeneous, finding consistent knowledge to share becomes increasingly challenging, and negative transfer~\citep{cui2022collaboration} may also arise. However, learning-to-guide knowledge is generic, making it easy for \ld to aggregate and distribute this knowledge in diverse scenarios, benefiting all clients. 

\subsubsection{More Clients}

In addition to experimenting with a total of 20 clients, we extend our evaluation by incorporating more clients created using the given Cifar100 dataset. With an increase in the number of clients, maintaining a consistent total data amount across all clients results in less local data on each client. In these scenarios, with a partial client participation ratio of $\rho=0.5$, our \ldl and \ldf can still maintain their superiority, as shown in \cref{tab:model}. Besides, \ld continues to outperform all baselines, demonstrating its robustness and scalability to a lower $\rho$.


\begin{table}[ht]
  \centering
  \caption{The communication and computation cost on Cifar100 in the default Dirichlet setting using HtFE$_8$. ``MB'' and ``s'' are short for megabyte and second, respectively. The time in ``()'' represents the cost of the warm-up period, several times less than local training. }
  \resizebox{!}{!}{
    \begin{tabular}{l|*{2}{c}|*{2}{c}}
    \toprule
    Items & \multicolumn{2}{c|}{Comm. (MB)} & \multicolumn{2}{c}{Computation (s)} \\
    \midrule
     & Up. & Down. & Client & Server \\
    \midrule
    LG-FedAvg & 3.93 & 3.93 & 6.18 & 0.04 \\
    FedGH & 1.75 & 3.93 & 9.53 & 0.37 \\
    FML & 70.57 & 70.57 & 8.63 & 0.07 \\
    FedKD & 63.02 & 63.02 & 9.04 & 0.07 \\
    FedMRL & 70.57 & 70.57 & 9.14 & 0.07 \\
    FedDistill & 0.34 & 0.76 & 6.52 & 0.03 \\
    FedProto & 1.75 & 3.89 & 6.65 & 0.04 \\
    \midrule
    \ldl & 0.34 & 0.76 & 7.49 (2.23) & 0.03 \\
    \ldf & 1.75 & 3.89 & 8.84 (2.24) & 0.04 \\
    \bottomrule
    \end{tabular}}
    \label{tab:three}
\end{table}



\subsubsection{Communication Cost} 

We consider both the upload and download bytes (across all participating clients) as part of the communication overhead in each iteration, using a float32 (= 4 bytes) data type in PyTorch~\citep{paszke2019pytorch} to store each floating number. In \cref{tab:three}, despite FML, FedKD, and FedMRL transmitting a relatively small global model, their communication costs remain significantly high compared to other methods that share lightweight components. The SVD technique in FedKD~\citep{wu2022communication}, does not significantly reduce the communication overhead. Given that we only upload the gradients of guiding vectors on the client, the communication cost of \ldl and \ldf is equivalent to that of FedDistill and FedProto, respectively. This cost falls within the lowest group among these methods. 

\subsubsection{Computation Cost} 

To capture essential operations, we measure the averaged GPU execution time of each client and the server on an idle GPU card in each iteration and show the time cost in \cref{tab:three}. 
As FedGH gathers prototypes after local training, it costs extra time for inferencing across the entire training set using the trained client model. In contrast, FedDistill and FedProto collect prototypical logits and features, respectively, concurrently with model training in each batch, thereby eliminating this additional cost. Besides, FedGH trains the global head on the server consuming relatively more power, even with one server epoch per iteration. Since we only average gradients on the server and update $\mathcal{G}$ once without backpropagation, our \ldl and \ldf demonstrate similar time-efficiency to FedDistill and FedProto, respectively. Due to the extra learning-to-guide process, \ld costs more client time than FedDistill and FedProto. However, \ldl still requires less time than FML, FedKD, FedMRL, and FedGH, and the improved test accuracy justifies this cost. 



\subsection{Properties of \ld}
\label{sec:property}

\subsubsection{\ld Prioritizes the Original Task}

\begin{figure}
	\centering
	\includegraphics[width=\linewidth]{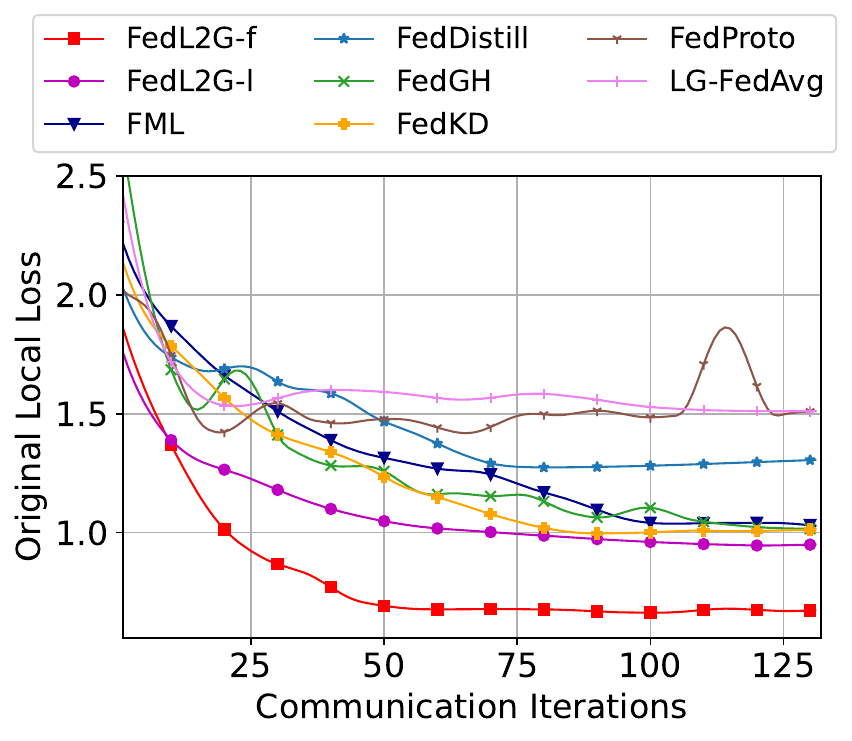}
	\caption{The averaged original local loss ($\ell_{ce}$) of all clients for HtFL methods on Cifar100 in the default Dirichlet setting using HtFE$_9$. }
	\label{fig:loss}
\end{figure}

Beyond presenting the test accuracy, we examine the training losses by examining the intrinsic training process. For each method, we illustrate only the original local loss, \ie, $\ell_{ce}$, in \cref{fig:loss}. 
These original local loss curves closely align with the accuracy trends in \cref{tab:model} (HtFE$_9$), indicating that lower original local loss corresponds to higher test accuracy in our scenarios. 
Since our \ld learns guiding vectors that help the client model focus more on its original task, \ldl and \ldf achieve the second-lowest and lowest losses, respectively. 

Besides the magnitude of the original local losses, our \ld method also offers advantages in smoothness and convergence speed. From \cref{fig:loss}, we observe that the loss curves of FedDistill, FedGH, FedProto, and LG-FedAvg fluctuate significantly in the beginning. The growth of $\ell_{ce}$ can be attributed to the mismatch of the shared global information and clients' tasks. 
Given that \ld focuses on clients' original tasks, we introduce more client-required information for guiding vectors, leading to a stable reduction in the original local loss. Because of the same benefits, \ldf can converge at a relatively early iteration and achieve the highest test accuracy simultaneously. Despite the lesser amount of guiding information in \ldl compared to \ldf, \ldl also demonstrates superiority in terms of smoothness and convergence when compared to FedDistill. 

\subsubsection{\ld Protects Private Information} 

\begin{figure}
	\centering
	\subfigure[\ldl.]{\includegraphics[width=0.48\linewidth]{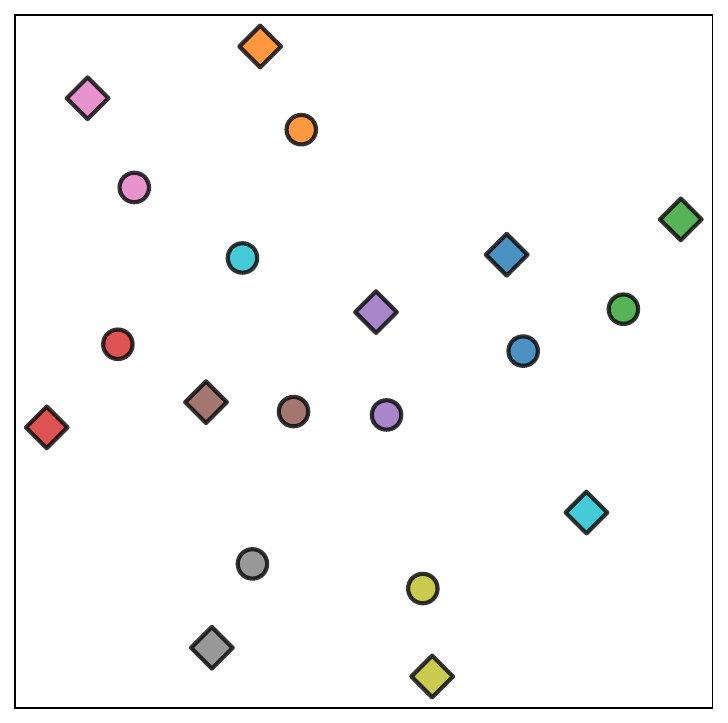}}
    \hfill
	\subfigure[\ldf.]{\includegraphics[width=0.48\linewidth]{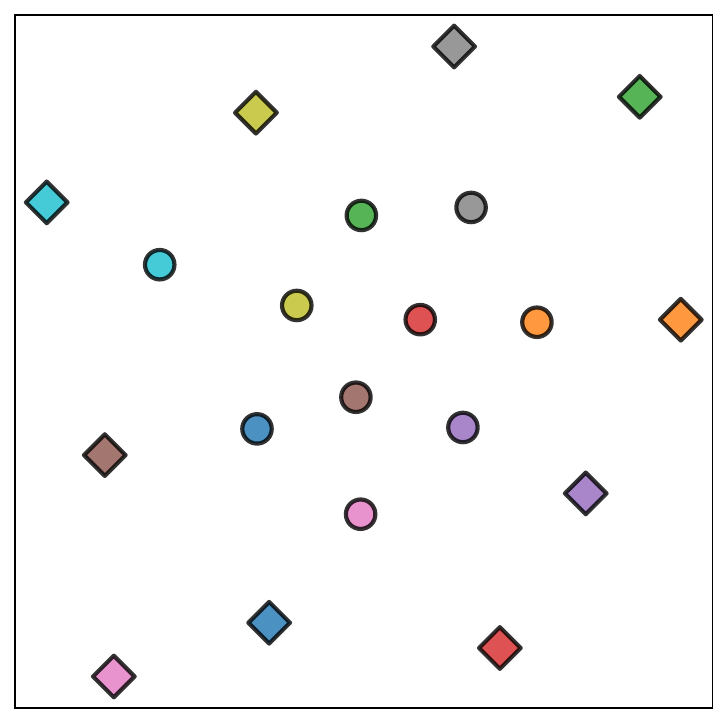}}
	\caption{The t-SNE visualization of guiding vectors (diamonds) and feature vectors (circles) on Cifar10 in the default Dirichlet setting using HtFE$_8$. Different colors represent different classes. \textit{Best viewed in color.}}
	\label{fig:tsne}
\end{figure}

Differing from FedDistill and FedProto, which gather data-derived prototypical logits and features from the clients, we collect the gradients of randomly initialized guiding vectors. These gradients are calculated using a complex formula (refer to \cref{eq:grad}) to reduce the original local losses for all clients. Therefore, our \ld does not directly upload client data-related information and safeguards the private feature information for clients. In a sense, logit vectors are also feature vectors with lower dimensions. Here, we illustrate the t-SNE~\citep{van2008visualizing} visualization of the global prototypes $\{{\bm g}^y\}_{y=1}^C$ and the guiding vectors $\{{\bm v}^y\}_{y=1}^C$ from \ldl and \ldf. As per \cref{fig:tsne}, guiding vectors differ greatly from global prototypes, which contain private information. 
This phenomenon is more pronounced in \ldf, where the distances between guiding vectors and global prototypes are larger than in \ldl, because the guiding vectors in \ldf have relatively more parameters and knowledge to learn. Given that a larger distance signifies improved discrimination and guidance for the class-wise vectors utilized in a guiding loss~\citep{zhang2024fedtgp}, our guiding vectors exhibit greater separability than the global prototypes, indicating enhanced guidance capability for the client models. 


\section{Conclusion}

We observe the original local loss growth phenomenon on the client in prior prototype-based HtFL methods when guided by global prototypes. Then we attribute this problem to the mismatch between the guiding objective and the client's original local objective. To address this issue, we propose a \ld approach to reduce the client's original objective when using guiding vectors by prioritizing the local objective during the learning of guiding vectors. The superiority of \ld is evidenced through theoretical analysis and extensive experiments. 

\section{Limitation}

While \ld effectively addresses the objective mismatch in HtFL and shows strong performance across various data and model heterogeneous scenarios, future work could extend \ld to handle more complex scenarios, such as clients with unstable connections. Despite these limitations, \ld significantly improves alignment between local and guiding objectives, enhancing HtFL efficiency.



\bibliography{main}
\bibliographystyle{icml2025}

\newpage
\appendix
\onecolumn

\setcounter{theorem}{0}
\setcounter{corollary}{0}
\setcounter{definition}{0}



\section{Algorithms}

Here is a detailed algorithm of our \ldl. 
Extending \cref{algo} to \ldf only requires notation substitutions. 

\vspace{-10pt}
\begin{algorithm}[ht]
	\caption{The Learning Processes in \ldl}
	\begin{algorithmic}[1]
		\Require 
		$N$ clients; 
		initial parameters ${\bm \theta}^0_1, \ldots, {\bm \theta}^0_N$ and $\mathcal{G}^0 = \{{\bm v}^{y, 0}\}_{y=1}^C$; 
		$\eta_c$: local learning rate; 
		$\eta_s$: server learning rate; 
		$\rho$: client joining ratio; 
		$E$: local epochs; 
		$T$: total communication iterations. 
		\Ensure 
        Well-trained client model parameters ${\bm \theta}_1, \ldots, {\bm \theta}_N$. 
        \State All clients split their training data into a study set $\mathcal{D}^s$ and a batch of quiz set $\mathcal{D}^q$. 
		\For{communication iteration $t=1, \ldots, T$}
		    \State Server samples a client subset $\mathcal{I}^t$ based on $\rho$.
		    \State Server sends $\mathcal{G}^{t-1}$ to each client in $\mathcal{I}^t$.
		    \For{Client $i \in \mathcal{I}^t$ in parallel}
                \If{$t > T'$}
                    \State Updates ${\bm \theta}^{t-1}_i$ to ${\bm \theta}^{t}_i$ using SGD for $E$ epochs via 
                    \Statex \qquad \qquad \qquad $\min_{{\bm \theta}_i} \ \mathbb{E}_{({\bm x}, y)\sim \mathcal{D}^s_i} [\ell_{ce}(f_i({\bm x}, {\bm \theta}^{t-1}_i), y) + \ell_{g}(f_i({\bm x}, {\bm \theta}^{t-1}_i), {\bm v}^{y, t-1})]$
                \Else
                    \State Marks ${\bm \theta}^{t-1}_i$ as ${\bm \theta}^{t}_i$. 
                \EndIf
                \State Executes a \textit{pseudo-train} step on a randomly sampled batch $\mathcal{B}^s_i$ via \cref{eq:pseudo-train} with ${\bm \theta}^{t}_i$. 
                \State Computes the gradients of $\mathcal{G}^{t-1}$, \ie, $\pi^t_i$, on $\mathcal{D}^q_i$ via \cref{eq:grad}.
		        \State Sends non-zero vectors among $\pi^t_i$ to the server. 
		    \EndFor
		    \State Server averages the non-zero vectors of $\pi^t_i, i \in \mathcal{I}^t$ for each class to obtain $\pi^t$. 
		    \State Server updates $\mathcal{G}^{t-1}$ to $\mathcal{G}^{t}$ via $\mathcal{G}^{t} = \mathcal{G}^{t-1} - \eta_s \pi^t$. 
		\EndFor
		\\
		\Return ${\bm \theta}^T_1, \ldots, {\bm \theta}^T_N$. 
	\end{algorithmic}
	\label{algo}
\end{algorithm}
\vspace{-10pt}

\section{Additional Experiments}
\label{sec:addexp}

\subsection{Additional Experimental Details}

\noindent\textbf{Datasets and environment. \ } 
We use four datasets with their respective download links: Cifar10\footnote{\url{https://pytorch.org/vision/main/generated/torchvision.datasets.CIFAR10.html}}, Cifar100\footnote{\url{https://pytorch.org/vision/stable/generated/torchvision.datasets.CIFAR100.html}}, Flowers102\footnote{\url{https://pytorch.org/vision/stable/generated/torchvision.datasets.Flowers102.html}}, and Tiny-ImageNet\footnote{\url{http://cs231n.stanford.edu/tiny-imagenet-200.zip}}. 
All our experiments are conducted on a machine with 64 Intel(R) Xeon(R) Platinum 8362 CPUs, 256G memory, eight NVIDIA 3090 GPUs, and Ubuntu 20.04.4 LTS. Most of our experiments can be completed within 48 hours, while others, involving many clients and extensive local training epochs, may take up to a week to finish. 

\noindent\textbf{Hyperparameter settings. \ } 
For our baseline methods, we set their hyperparameters following existing work~\citep{zhang2024fedtgp, zhang2024upload}. As for our \ldl and \ldf, we tune the server learning rate $\eta_s$ and the number of warm-up rounds $T'$ by grid search on the Cifar100 dataset in the default Dirichlet setting with HtFE$_8$ and use an identical setting on all experimental tasks without further tuning. Specifically, We search for $\eta_s$ in the range $\{0.01, 0.05, 0.1, 0.5, 1, 10, 50, 100, 500\}$ and for $T'$ in the range $\{0, 1, 10, 20, 50, 100\}$. We set $T' = 50$ for all scenarios, with the warm-up cost considered negligible since no local training is performed. 
We set $\eta_s=0.1$ for \ldl and set $\eta_s=100$ for \ldf. The $\eta_s$ hyperparameters of \ldl and \ldf differ due to their discrepancy in the learnable knowledge capacity of the guiding vectors. The dimension of the guiding vectors in \ldf is larger than in \ldl, necessitating more server updates. 

\noindent\textbf{The small auxiliary model for FML, FedKD, and FedMRL. \ } 
As FML, FedKD, and FedMRL utilize a global auxiliary model for mutual distillation, this auxiliary model needs to be as compact as possible to minimize communication overhead during model parameter transmission~\citep{wu2022communication}. Therefore, we opt for the smallest model within each group of heterogeneous models to serve as the auxiliary model in all scenarios.

\subsection{Hyperparameter Study}
\label{sec:hyper}

We conduct a hyperparameter study here to study the influence of two hyperparameters: the server learning rate $\eta_s$ and the number of warm-up rounds $T'$. 
\begin{itemize}
\item \textbf{$\eta_s$. \ } From \cref{tab:eta_s}, we know that \ldl and \ldf benefit from distinct ranges of $\eta_s$, also attributed to their different trainable parameters and learning capacities. Moreover, \ldf demonstrates higher optimal accuracy than \ldl, while \ldl yields a more stable outcome across different $\eta_s$. 

\item \textbf{$T'$. \ } The warm-up phase, which includes steps \textbf{\cir{1}}, \textbf{\cir{3}}, \textbf{\cir{4}}, \textbf{\cir{5}}, \textbf{\cir{6}}, \textbf{\cir{7}}, is computationally lightweight and closely mirrors the main FL process, with the exception that step \textbf{\cir{2}} (local model updates) is skipped. This design ensures that the warm-up phase requires minimal additional effort. Besides, we only require participating clients to join instead of all clients in the warm-up phase. As shown in \cref{tab:T'}, \ld maintains competitive performance even with no warm-up ($T'=0$). The introduction of the warm-up phase does not impact the overall convergence speed. However, a short warm-up phase enhances guiding vector initialization, improving subsequent rounds' performance. Notably, \ldf benefits more from a warm-up phase due to the higher learning capacity of the high-dimensional feature space. Overly large $T'$ negatively impacts both variants due to overfitting on untrained client models.
\end{itemize}

\begin{table*}[ht]
    \centering
    \caption{The test accuracy (\%) of \ldl and \ldf on Cifar100 in the default Dirichlet setting using HtFE$_8$ with different $\eta_s$. }
    \resizebox{!}{!}{
    \begin{tabular}{l|ccccccc}
    \toprule
     & $\eta_s=0.01$ & $\eta_s=0.05$ & $\eta_s=0.1$ & $\eta_s=0.5$ & $\eta_s=1$ \\
    \midrule
    \ldl & 41.7 & 41.6 & \textbf{42.3} & 41.6 & 41.8 \\
    \hline\hline
     & $\eta_s=1$ & $\eta_s=10$ & $\eta_s=50$ & $\eta_s=100$ & $\eta_s=500$ \\
    \midrule
    \ldf & 41.1 & 42.0 & 43.5 & \textbf{43.8} & 41.4 \\
    \bottomrule
    \end{tabular}}
    \label{tab:eta_s}
\end{table*}

\begin{table*}[ht]
    \centering
    \caption{The test accuracy (\%) of \ldl and \ldf on Cifar100 in the default Dirichlet setting using HtFE$_8$ with different $T'$. The results in ``()'' represent ``the total number of converged rounds including the warm-up round''.}
    \resizebox{!}{!}{
    \begin{tabular}{l|ccccccc}
    \toprule
     & $T'=0$ (no warming-up) & $T'=1$ & $T'=10$ & $T'=20$ & $T'=50$ & $T'=100$ \\
    \midrule
    \ldl & 41.7 (160) & 41.8 (156) & 41.7 (165) & 42.0 (158) & \textbf{42.3 (159)} & 41.8 (161) \\
    \ldf & 40.9 (163) & 41.6 (160) & 43.0 (155) & 43.6 (157) & \textbf{43.8 (160)} & 43.6 (162) \\
    \bottomrule
    \end{tabular}}
    \label{tab:T'}
\end{table*}

\subsection{Different Quiz Set Size} 

\begin{table*}[ht]
    \centering
    \caption{Test accuracy (\%) on Cifar100 in the Dirichlet setting using HtFE$_8$ with different quiz set size (qss).}
    \resizebox{!}{!}{
    \begin{tabular}{l|ccccccc}
    \toprule
     & qss=10 (original) & qss=2 & qss=5 \\
    \midrule
    \ldl & \textbf{42.3} & 42.2 & \textbf{42.3} \\
    \ldf & 43.8 & \textbf{44.2} & 43.4 \\
    \bottomrule
    \end{tabular}}
    \label{tab:qss}
\end{table*}

The quiz set size (qss) is not a hyperparameter, as it originally matches the training batch size, which is consistent across all baselines. To explore its sensitivity, we vary qss and present the results in \cref{tab:qss}. The quiz set is not an additional dataset but a small portion held out from the original training data, ensuring fairness and no extra data advantage over other baselines. As shown in \cref{tab:qss}, only 2 to 5 samples are sufficient for our \ld to achieve strong performance. This implementation is straightforward and supported by tools like higher\footnote{\url{https://github.com/facebookresearch/higher}}~\citep{grefenstette2019generalized}.

\subsection{Effectiveness of Server Aggregation}

\begin{table*}[ht]
    \centering
    \caption{Test accuracy (\%) on two datasets in the Dirichlet setting using HtFE$_8$.}
    \resizebox{!}{!}{
    \begin{tabular}{l|ccccccc}
    \toprule
    & Cifar10 & Cifar100 \\
    \midrule
    Local Training & 83.2 & 35.6 \\
    \ldl & 86.5 & 42.3 \\
    \ldf & \textbf{87.6} & \textbf{43.8} \\
    \bottomrule
    \end{tabular}}
    \label{tab:avg}
\end{table*}

Averaging, as introduced in FedProto~\citep{tan2022fedproto}, is a widely accepted and effective practice in FL for aggregating and sharing global information under both data and model heterogeneity. In our \ld framework, updating local models using global guiding vectors plays a crucial role in aligning local models and promoting consistency in their feature extraction. Without the global guiding vectors, local models lack this critical alignment, resulting in significantly poorer performance, as demonstrated in \cref{tab:avg}.

\subsection{Privacy Discussion}

\begin{table*}[ht]
    \centering
    \caption{Test accuracy (\%) on Cifar100 in the Dirichlet setting using HtFE$_8$ with a noise scale of $s$ and perturbation coefficient $p$.}
    \resizebox{!}{!}{
    \begin{tabular}{l|ccccccc}
    \toprule
    & Original & Add Noise ($s=0.05$, $p=0.1$) & Add Noise ($s=0.05$, $p=0.2$) \\
    \midrule
    \ldl & 42.3 & 41.8 & 41.7 \\
    \ldf & 43.8 & 42.9 & 42.6 \\
    \bottomrule
    \end{tabular}}
    \label{tab:noise}
\end{table*}

As mentioned in \cref{sec:property}, our \ld method does not upload raw features or local class prototypes. Instead, it uploads gradients of guiding vectors, as shown in \textit{Line 12} of \cref{algo}, which are initialized randomly and iteratively refined through client feedback. These gradients are not directly related to sensitive local data or class-specific statistical information. \cref{fig:tsne} demonstrates that guiding vectors differ significantly from class prototypes, ensuring privacy. To further enhance privacy, we incorporated Gaussian noise into the gradients of guiding vectors following the approach in~\citep{tan2022federated}. \ld retains strong performance while improving privacy protection (see \cref{tab:noise}).

\subsection{More Local Training Epochs}

\begin{table*}[ht]
  \centering
  \caption{The test accuracy (\%) on Cifar100 in the default Dirichlet setting using HtFE$_8$ with different local training epochs. }
  \resizebox{!}{!}{
    \begin{tabular}{l|*{3}{c}}
    \toprule
     & $E=5$ & $E=10$ & $E=20$ \\
    \midrule
    LG-FedAvg & 40.3$\pm$.2 & 40.5$\pm$.1 & 40.9$\pm$.2 \\
    FedGH & 41.1$\pm$.3 & 39.9$\pm$.3 & 40.2$\pm$.4 \\
    FML & 39.1$\pm$.3 & 38.0$\pm$.2 & 36.0$\pm$.2 \\
    FedKD & 41.1$\pm$.1 & 40.4$\pm$.2 & 39.1$\pm$.3 \\
    FedMRL & 42.1$\pm$.8 & 42.4$\pm$.7 & 42.9$\pm$.6 \\
    FedDistill & 41.0$\pm$.3 & 41.3$\pm$.2 & 41.1$\pm$.4 \\
    FedProto & 38.0$\pm$.5 & 38.1$\pm$.4 & 38.7$\pm$.5 \\
    \midrule
    \ldl & 42.2$\pm$.2 & 42.0$\pm$.2 & 42.1$\pm$.1 \\
    \ldf & \textbf{43.7$\pm$.1} & \textbf{43.8$\pm$.2} & \textbf{44.3$\pm$.3} \\
    \bottomrule
    \end{tabular}}
    \label{tab:epoch}
\end{table*}

Increasing the number of local epochs, denoted by $E$, in each communication iteration can reduce the total number of iterations required for convergence, consequently lowering total communication overhead~\citep{mcmahan2017communication, zhang2024upload}. In \cref{tab:epoch}, FedGH experiences approximately a 1\% decrease in accuracy when $E \ge 10$. Since the globally shared model struggles with data heterogeneity, FML and FedKD also exhibit performance degradation with a larger $E$, albeit more severe. Specifically, FML and FedKD continue to decrease from $E=5$ to $E=20$, with FML dropping by 3.1\% and FedKD dropping by 2.0\%. In contrast, our \ldl and \ldf consistently uphold superior performance even with a larger $E$. Remarkably, \ldf shows an increase of 0.6\% in accuracy from $E=5$ to $E=20$, showcasing its exceptional adaptability in scenarios with low communication quality. 

\subsection{Additional Data Heterogeneous Degrees}

\begin{table*}[ht]
    \centering
    \caption{Test accuracy (\%) on Cifar100 in the Dirichlet setting using HtFE$_8$ with varying $\beta$. The results in ``()'' mean the total number of converged rounds including the warm-up phase for \ld.}
    \resizebox{!}{!}{
    \begin{tabular}{l|ccccccc}
    \toprule
     & \(\beta = 0.01\) & \(\beta = 0.1\) & \(\beta = 0.5\) & \(\beta = 1\) \\
    \midrule
    LG-FedAvg & 66.6 (178) & 40.7 (190) & 21.3 (273) & 15.7 (141) \\
    FedGH & 65.2 (146) & 41.0 (226) & 21.2 (232) & 15.5 (184) \\
    FML & 64.5 (370) & 39.9 (287) & 20.0 (150) & 16.0 (318) \\
    FedKD & 64.9 (285) & 40.6 (198) & 21.5 (166) & 16.3 (288) \\
    FedMRL & 68.8 (181) & 41.2 (170) &  22.3 (152) & 16.3 (567)  \\
    FedDistill & 67.0 (338) & 41.5 (216) & 22.1 (161) & 16.4 (273) \\
    FedProto & 60.6 (540) & 36.3 (533) & 18.3 (570) & 12.6 (369) \\
    \ldl & 68.2 (196) & 42.3 (176) & 22.1 (189) & 16.7 (172) \\
    \ldf & \textbf{70.6 (257)} & \textbf{43.8 (235)} & \textbf{23.3 (225)} & \textbf{16.8 (210)} \\
    \bottomrule
    \end{tabular}}
    \label{tab:beta}
\end{table*}

In \cref{sec:perform}, we have evaluated \ld under three levels of data heterogeneity: pathological, Dirichlet ($\beta=0.1$), and Dirichlet ($\beta=0.01$). These are standard settings for studying data heterogeneity~\citep{zhang2022fedala}. To further study our \ld's robustness to various data heterogeneity, we conducted additional experiments using $\beta$ values of 0.01, 0.5, and 1. \cref{tab:beta} demonstrates that \ld consistently outperforms baselines across all settings, even as data heterogeneity varies. While larger $\beta$ results in less skewed data distributions, it reduces per-class data availability for clients, impacting overall performance. The communication efficiency remains consistent across different scenarios, as the gradients of the guiding vectors retain the same shape in every communication round. 
Although FedMRL performs sub-optimally compared to other baselines, its convergence rate varies significantly, ranging from 152 to 567 iterations, whereas our \ld demonstrates stable and consistent convergence rates. Besides, FedProto requires much more iterations to converge. 

\subsection{Feature Shift Scenario}

\begin{table*}[ht]
    \centering
    \caption{Test accuracy (\%) on DomainNet in the feature shift scenario using HtFE$_4$}
    \resizebox{!}{!}{
    \begin{tabular}{l|ccccccc}
    \toprule
     & DomainNet \\
    \midrule
    LG-FedAvg & 26.9 \\
    FedGH & 25.0 \\
    FML & 24.9 \\
    FedKD & 25.0 \\
    FedMRL & 24.4 \\
    FedDistill & 26.8 \\
    FedProto & 21.2 \\
    \midrule
    \ldl & 27.3 \\
    \ldf & \textbf{28.2} \\
    \bottomrule
    \end{tabular}}
    \label{tab:domainnet}
\end{table*}

We have demonstrated the superiority of our \ld in the main body. Here, we further evaluate its effectiveness in a new data heterogeneity scenario involving feature shift~\cite{li2022federated}, where each client has access to all labels but varies in data features (\eg, image styles). This scenario is commonly simulated using the DomainNet dataset~\cite{peng2019moment}, which presents a challenging task~\cite{yang2025feature}. Specifically, we assign each client a subset of DomainNet from distinct domains. The excellent performance of \ld in \cref{tab:domainnet} further validates our \ld's applicability and robustness. We also observe that FedMRL performs worse than most other baselines on DomainNet, despite achieving high accuracy in the main experiments.

\section{Theoretical Analysis}
\label{sec:theo}
\setcounter{equation}{0}
\setcounter{assumption}{0}
\numberwithin{equation}{section}

Here we bring some existing equations for convenience. 
Recall that we have $N$ clients training their heterogeneous local models (with parameters ${\bm \theta}_1, \ldots, {\bm \theta}_N$) using their private and heterogeneous training data $\mathcal{D}_1, \ldots, \mathcal{D}_N$. Besides, they share global guiding vectors $\mathcal{G} = \{{\bm v}^y\}_{y=1}^C$, with the assistance of a server to facilitate collaborative learning. Formally, the objective of \ld is
\begin{equation}
    \min_{{\bm \theta}_1, \ldots, {\bm \theta}_N} \ \sum_{i=1}^N \frac{|\mathcal{D}_i|}{D} \mathcal{L}_{\mathcal{D}_i}({\bm \theta}_i, \mathcal{G}), \label{eq:1}
\end{equation}
where the total client loss $\mathcal{L}_{\mathcal{D}_i}$ is defined by
\begin{equation}
    \mathcal{L}_{\mathcal{D}_i}({\bm \theta}_i, \mathcal{G}) := \mathbb{E}_{({\bm x}, y)\sim \mathcal{D}_i} [\ell_{ce}(f_i({\bm x}, {\bm \theta}_i), y) + \ell_{g}(f_i({\bm x}, {\bm \theta}_i), {\bm v}^y)], \label{eq:2}
\end{equation}
and the original local loss $\mathcal{L}'_{\mathcal{D}_i}$ is defined by 
\begin{equation}
    \mathcal{L}'_{\mathcal{D}_i}({\bm \theta}_i, \mathcal{G}) := \mathbb{E}_{({\bm x}, y)\sim \mathcal{D}_i} [\ell_{ce}(f_i({\bm x}, {\bm \theta}_i), y)]. \label{eq:0}
\end{equation}
Here we consider $\ldl$ for simplicity, and it is easy to extend theoretical analysis to $\ldf$ by substituting $\ell_{g}(f_i({\bm x}, {\bm \theta}_i), {\bm v}^y)$ with $\ell_{g}(h_i({\bm x}, {\bm \theta}_i^h), {\bm v}^y)$. 
We optimize global $\mathcal{G}$ by 
\begin{equation}
    \mathcal{G}^t = \mathcal{G}^{t-1} - \eta_s \frac{1}{N} \sum_{i\in [N]} \nabla_{\mathcal{G}^{t-1}} \mathbb{E}_{({\bm x}, y)\sim \mathcal{D}^q_i} [\ell_{ce}(f_i({\bm x}, {\bm \theta}_i - \eta_c \nabla_{{\bm \theta}_i} \mathcal{L}_{\mathcal{B}^s_i}({\bm \theta}_i, \mathcal{G}^{t-1})), y)], \label{eq:3}
\end{equation}
where we consider full participation for simplicity. 
The convergence analysis of HtFL typically considers an arbitrary client, incorporating global information (\eg, $\mathcal{G}$) to facilitate collaboration~\cite{tan2022fedproto, yi2024federated}.
Thus, in the following, we omit the client notation $i$ and some corresponding notations, such as $\mathcal{D}_i$, for simplicity. 

To further examine the local training process, in addition to the communication iteration notation $t$, we introduce $e \in \{1/2,1,2,\ldots,E\}$ to represent the local step. We denote the $e$th local training step in iteration $t$ as $tE+e$. Specifically, $tE+1/2$ refers to the moment when clients receive $\mathcal{G}$ before local training. We adopt four assumptions, partially based on FedProto~\citep{tan2022fedproto}. 

\begin{assumption}[Unbiased Gradient and Bounded Variance]
    The stochastic gradient $\omega^t = \nabla \mathcal{L}_{\xi}({\bm \theta}^t, \mathcal{G}^t)$ is an unbiased estimation of each client's gradient w.r.t. its loss: 
    $$\mathbb{E}_{\xi \sim \mathcal{D}} [\omega^t] = \nabla \mathcal{L}({\bm \theta}^t, \mathcal{G}) = \nabla \mathcal{L}^t.$$ and its variance is bounded by $\sigma^2$: 
    $$\mathbb{E}[||\omega^t - \nabla \mathcal{L}^t||_2^2] \le \sigma^2.$$\label{ass:1}
\end{assumption}
\begin{assumption}[Bounded Gradient]
    The expectation of the stochastic gradient $\omega^t$ and $\omega'^t = \nabla \mathcal{L}'_{\xi}({\bm \theta}^t, \mathcal{G}^t)$ are bounded by $R$ and $R'$, respectively: 
    $$\mathbb{E} [||\omega^t||_2] \le R, \quad \mathbb{E} [||\omega'^t||_2] \le R'.$$ \label{ass:2}
\end{assumption}
\begin{assumption}[Lipschitz Smoothness]
    Each total local objective $\mathcal{L}$ is $L_1$-Lipschitz smooth, which also means the gradient of $\mathcal{L}$ is $L_1$-Lipschitz continuous, i.e., 
    $$||\nabla \mathcal{L}^{t_1} - \nabla \mathcal{L}^{t_2}||_2 \le L_1||{\bm \theta}^{t_1} - {\bm \theta}^{t_2}||_2, \quad \forall t_1, t_2 > 0,$$ 
    which implies the following quadratic bound, 
    $$\mathcal{L}^{t_1} - \mathcal{L}^{t_2} \le \left<\nabla \mathcal{L}^{t_2}, ({\bm \theta}^{t_1} - {\bm \theta}^{t_2}) \right> + \frac{1}{2} L_1 ||{\bm \theta}^{t_1} - {\bm \theta}^{t_2}||_2^2, \quad \forall t_1, t_2 > 0.$$
    \label{ass:3}
\end{assumption}

Given \cref{ass:1} and \cref{ass:2}, any client's gradient \wrt $\mathcal{G}$ is 
\begin{align}
    \pi^{t-1} &= \nabla_{\mathcal{G}^{t-1}} \mathbb{E}_{({\bm x}, y)\sim \mathcal{D}^q} [\ell_{ce}(f({\bm x}, {\bm \theta} - \eta_c \nabla_{{\bm \theta}} \mathcal{L}_{\mathcal{B}^s}({\bm \theta}, \mathcal{G}^{t-1})), y)] \\
    &= \mathbb{E}_{({\bm x}, y)\sim \mathcal{D}^q} [\nabla_{\mathcal{G}^{t-1}} \ell_{ce}(f({\bm x}, {\bm \theta} - \eta_c \nabla_{{\bm \theta}} \mathcal{L}_{\mathcal{B}^s}({\bm \theta}, \mathcal{G}^{t-1})), y)] \\
    &= \mathbb{E}_{({\bm x}, y)\sim \mathcal{D}^q} [\nabla_1 \ell_{ce} \cdot \nabla_2 f \cdot \nabla_{\mathcal{G}^{t-1}}({\bm \theta} - \eta_c \nabla_{{\bm \theta}} \mathcal{L}_{\mathcal{B}^s}({\bm \theta}, \mathcal{G}^{t-1}))] \\
    &= - \eta_c \mathbb{E}_{({\bm x}, y)\sim \mathcal{D}^q} [\nabla_1 \ell_{ce} \cdot \nabla_2 f \cdot \nabla_{\mathcal{G}^{t-1}} \nabla_{{\bm \theta}} \mathcal{L}_{\mathcal{B}^s}({\bm \theta}, \mathcal{G}^{t-1})] \\
    &= - \eta_c \mathbb{E}_{({\bm x}, y)\sim \mathcal{D}^q} \{\nabla_1 \ell_{ce} \cdot \nabla_2 f \cdot \mathbb{E}_{({\bm x}', y')\sim \mathcal{B}^s} [\nabla_{\mathcal{G}^{t-1}} \nabla_{{\bm \theta}} \ell_{g}(f({\bm x}', {\bm \theta}), {\bm v}^{y'})]\} \\
    &= - \eta_c \mathbb{E}_{({\bm x}, y)\sim \mathcal{D}^q} \{\nabla_1 \ell_{ce} \cdot \nabla_2 f \cdot \mathbb{E}_{({\bm x}', y')\sim \mathcal{B}^s} [\nabla_2 f \cdot \nabla_{\mathcal{G}^{t-1}} \nabla_1 \ell_{g}]\} \\
    &= 2 \eta_c \mathbb{E}_{({\bm x}, y)\sim \mathcal{D}^q} \{\nabla_1 \ell_{ce} \cdot \nabla_2 f \cdot \mathbb{E}_{({\bm x}', y')\sim \mathcal{B}^s} [\nabla_2 f]\}, \label{eq:4}
\end{align}
where $\nabla_1 \ell_{ce} := \nabla_{a_1} \ell_{ce}(a_1, a_2)$, indicating the derivative of $\ell_{ce}$ \wrt the first variable, and so for $\nabla_2 f$ and $\nabla_1 \ell_{g}$. Under \cref{ass:1}, we can mimic regular training through the pseudo-train step \textbf{\cir{3}}, as $\mathcal{B}^s$ is randomly re-sampled in each iteration. All the derivatives in \cref{eq:4} are bounded under \cref{ass:2}. 

Then, we have two key lemmas:
\begin{lemma}
    Let \cref{ass:1} and \cref{ass:3} hold. The total client loss of an arbitrary client can be bounded: 
    $$\mathbb{E}[\mathcal{L}^{(t+1)E}] \le \mathcal{L}^{tE+1/2} + (\frac{L_1\eta_c^2}{2} - \eta_c) \sum_{e=1/2}^{E-1} ||\nabla \mathcal{L}^{tE+e}||_2^2 + \frac{L_1E\eta_c^2\sigma^2}{2}.$$ \label{lm:1}
\end{lemma}

\begin{proof}
    This lemma focuses solely on local training at the client level, incorporating both the original local objective and the guiding objective. It can be easily derived by substituting the relevant notations from Lemma 1 of the prototype-based HtFL method, FedProto. 
\end{proof}

\begin{lemma}
    Let \cref{ass:2} and \cref{ass:3} hold. After the guiding vectors are updated on the server and downloaded to clients, the total client loss of an arbitrary client can be bounded: 
    $$\mathbb{E}[\mathcal{L}^{(t+1)E+1/2}] \le \mathcal{L}^{(t+1)E} + 2\eta_c^2\eta_sR'ER.$$ \label{lm:2}
\end{lemma}

\begin{proof}
    \begin{align}
         \mathcal{L}^{(t+1)E+1/2} &= \mathcal{L}^{(t+1)E} + \mathcal{L}^{(t+1)E+1/2} - \mathcal{L}^{(t+1)E} \\
         &= \mathcal{L}^{(t+1)E} + ||f({\bm \theta}^{(t+1)E}) - \mathcal{G}^{(t+2)E}||_2 - ||f({\bm \theta}^{(t+1)E}) - \mathcal{G}^{(t+1)E}||_2 \\
         &\stackrel{(a)}{\le} \mathcal{L}^{(t+1)E} + ||\mathcal{G}^{(t+2)E} - \mathcal{G}^{(t+1)E}||_2 \\
         &= \mathcal{L}^{(t+1)E} + \eta_s||\mathbb{E}_{[N]} \pi^{(t+1)E}||_2 \\
         &\stackrel{(b)}{\le} \mathcal{L}^{(t+1)E} + \eta_s\mathbb{E}_{[N]} ||\pi^{(t+1)E}||_2 \\
         &\stackrel{(c)}{\le} \mathcal{L}^{(t+1)E} + 2\eta_c\eta_s\mathbb{E}_{[N]} \mathbb{E}_{\mathcal{D}} ||\nabla_1 \ell_{ce}^{(t+1)E} \cdot \nabla_2 f^{(t+1)E} \cdot \mathbb{E}_{\xi} [\nabla_2 f^{(t+1)E}]||_2 \\
         &\stackrel{(d)}{\le} \mathcal{L}^{(t+1)E} + 2\eta_c\eta_sR'\mathbb{E}_{[N]} ||\mathbb{E}_{\xi} \nabla_2 f^{(t+1)E}||_2 \\
         &\stackrel{(e)}{\le} \mathcal{L}^{(t+1)E} + 2\eta_c\eta_sR'\mathbb{E}_{[N]} \mathbb{E}_{\xi} ||\nabla_2 f^{(t+1)E}||_2 \\
         &= \mathcal{L}^{(t+1)E} + 2\eta_c\eta_sR'\mathbb{E}_{[N]} \mathbb{E}_{\xi} ||{\bm \theta}^{(t+1)E} - {\bm \theta}^{tE}||_2 \\
         &\stackrel{(f)}{\le} \mathcal{L}^{(t+1)E} + 2\eta_c^2\eta_sR'\mathbb{E}_{[N]} \mathbb{E}_{\xi} \sum_{e=1/2}^{E-1} ||\omega^{tE+e}||_2 
    \end{align}

    Take expectations of random variable $\xi$, we have 
    \begin{align}
        \mathbb{E}[\mathcal{L}^{(t+1)E+1/2}] &\le \mathcal{L}^{(t+1)E} + 2\eta_c^2\eta_sR'\mathbb{E}_{[N]} \mathbb{E}_{\xi} \sum_{e=1/2}^{E-1} ||\omega^{tE+e}||_2 \\
        &\stackrel{(g)}{\le} \mathcal{L}^{(t+1)E} + 2\eta_c^2\eta_sR'ER. 
    \end{align}

    In the above inequations, ($a$) follows from $||a-b||_2 - ||a-c||_2 \le ||b-c||_2$; ($b$), ($c$), ($e$), and ($f$) follow from $||\sum a_j||_2 \le \sum ||a_j||_2$, where $\mathbb{E}_{\mathcal{D}} a$ denotes taking expectations of $a$ over set $\mathcal{D}$, \eg, $\mathbb{E}_{[N]} a$ means $\mathbb{E}_{i \sim \{1, \ldots, N\}} a_j$; ($d$) follows from \cref{ass:1} and \cref{ass:2}, where $\nabla \mathcal{L}'({\bm \theta}, \mathcal{G}) = \nabla_1 \ell_{ce} \cdot \nabla_2 f$; ($g$) follows from \cref{ass:2}. 
\end{proof}

Then, we have 

\begin{theorem}[One-iteration deviation]
    Let \cref{ass:1} to \cref{ass:3} hold. For an arbitrary client, after every communication iteration (with $\mathcal{G}$ for collaboration), we have
    $$\mathbb{E}[\mathcal{L}^{(t+1)E+1/2}] \le \mathcal{L}^{tE+1/2} + (\frac{L_1\eta_c^2}{2} - \eta_c) \sum_{e=1/2}^{E-1} ||\nabla \mathcal{L}^{tE+e}||_2^2 + \frac{L_1E\eta_c^2\sigma^2}{2} + 2\eta_c^2\eta_sR'ER.$$ \label{th:c1}
\end{theorem}

\begin{proof}
    Taking expectation of ${\bm \theta}$ on both sides in \cref{lm:2}, we have 
    \begin{equation}
        \mathbb{E}[\mathcal{L}^{(t+1)E+1/2}] \le \mathbb{E}[\mathcal{L}^{(t+1)E}] + 2\eta_c^2\eta_sR'ER. \label{eq:T1}
    \end{equation}
    Then summing \cref{eq:T1} and \cref{lm:1} up, we have 
    \begin{equation}
        \mathbb{E}[\mathcal{L}^{(t+1)E+1/2}] \le \mathcal{L}^{tE+1/2} + (\frac{L_1\eta_c^2}{2} - \eta_c) \sum_{e=1/2}^{E-1} ||\nabla \mathcal{L}^{tE+e}||_2^2 + \frac{L_1E\eta_c^2\sigma^2}{2} + 2\eta_c^2\eta_sR'ER. \label{eq:T2}
    \end{equation}
\end{proof}

\begin{theorem}[Non-convex convergence rate of \ld]
    Let \cref{ass:1} to \cref{ass:3} hold and $\Delta = \mathcal{L}^0 - \mathcal{L}^*$, where $\mathcal{L}^*$ refers to the local optimum. Given \cref{th:c1}, for an arbitrary client and an arbitrary constant $\epsilon$, our \ld has a non-convex convergence rate $\mathcal{O}(1/T)$ with 
    $$\frac{1}{T} \sum_{t=0}^{T-1} \sum_{e=1/2}^{E-1} \mathbb{E}[||\nabla \mathcal{L}^{tE+e}||_2^2] \le \frac{\frac{2\Delta}{T} + L_1E\eta_c^2\sigma^2 + 4\eta_c^2\eta_sR'ER}{2\eta_c - L_1\eta_c^2} < \epsilon,$$
    where $0 < \eta_c < \frac{2\epsilon}{L_1(E\sigma^2 + \epsilon) + 4\eta_sR'ER}$ and $\eta_s > 0$. 
\end{theorem}

\begin{proof}
    By interchanging the left and right sides of \cref{eq:T2}, we can get 
    \begin{equation}
        \sum_{e=1/2}^{E-1} ||\nabla \mathcal{L}^{tE+e}||_2^2 \le \frac{\mathcal{L}^{tE+1/2} - \mathbb{E}[\mathcal{L}^{(t+1)E+1/2}] + \frac{L_1E\eta_c^2\sigma^2}{2} + 2\eta_c^2\eta_sR'ER}{\eta_c - \frac{L_1\eta_c^2}{2}},
    \end{equation}
    when $\eta_c - \frac{L_1\eta_c^2}{2} > 0$, \ie, $0 < \eta_c < \frac{2}{L_1}$. 
    Taking the expectation of ${\bm \theta}$ on both sides and summing all inequalities overall communication iterations, we obtain
    \begin{equation}
        \frac{1}{T} \sum_{t=0}^{T-1} \sum_{e=1/2}^{E-1} \mathbb{E}[||\nabla \mathcal{L}^{tE+e}||_2^2] \le \frac{\frac{1}{T} \sum_{t=0}^{T-1} (\mathcal{L}^{tE+1/2} - \mathbb{E}[\mathcal{L}^{(t+1)E+1/2}]) + \frac{L_1E\eta_c^2\sigma^2}{2} + 2\eta_c^2\eta_sR'ER}{\eta_c - \frac{L_1\eta_c^2}{2}}.
    \end{equation}
    Let $\Delta = \mathcal{L}^0 - \mathcal{L}^* > 0$, we have $\frac{1}{T} \sum_{t=0}^{T-1} (\mathcal{L}^{tE+1/2} - \mathbb{E}[\mathcal{L}^{(t+1)E+1/2}]) \le \Delta$ and 
    \begin{equation}
        \frac{1}{T} \sum_{t=0}^{T-1} \sum_{e=1/2}^{E-1} \mathbb{E}[||\nabla \mathcal{L}^{tE+e}||_2^2] \le \frac{\frac{2\Delta}{T} + L_1E\eta_c^2\sigma^2 + 4\eta_c^2\eta_sR'ER}{2\eta_c - L_1\eta_c^2}. \label{eq:1T}
    \end{equation}
    Given any $\epsilon > 0$, let 
    \begin{equation}
        \frac{\frac{2\Delta}{T} + L_1E\eta_c^2\sigma^2 + 4\eta_c^2\eta_sR'ER}{2\eta_c - L_1\eta_c^2} < \epsilon, 
    \end{equation}
    we have
    \begin{equation}
        T > \frac{2\Delta}{\epsilon\eta_c(2-L_1\eta_c) - \eta_c^2(L_1E\sigma^2 + 4\eta_sR'ER)}. 
    \end{equation}
    In this context, we have
    \begin{equation}
        \frac{1}{T} \sum_{t=0}^{T-1} \sum_{e=1/2}^{E-1} \mathbb{E}[||\nabla \mathcal{L}^{tE+e}||_2^2] \le \epsilon, 
    \end{equation}
    when 
    \begin{equation}
        0 < \eta_c < \frac{2\epsilon}{L_1(E\sigma^2 + \epsilon) + 4\eta_sR'ER} < \frac{2}{L_1}, 
    \end{equation}
    and 
    \begin{equation}
        \eta_s > 0
    \end{equation}

    Since all the notations of the right side in \cref{eq:1T} are given constants except for $T$, our \ld's non-convex convergence rate is $\epsilon \sim \mathcal{O}(1/T)$. 
\end{proof}

\section{Visualizations of Data Distributions}

We illustrate the data distributions on all clients in the above experiments in the following. 

\begin{figure*}[ht]
	\centering
	\hfill
	\subfigure[Cifar10]{\includegraphics[width=0.48\linewidth]{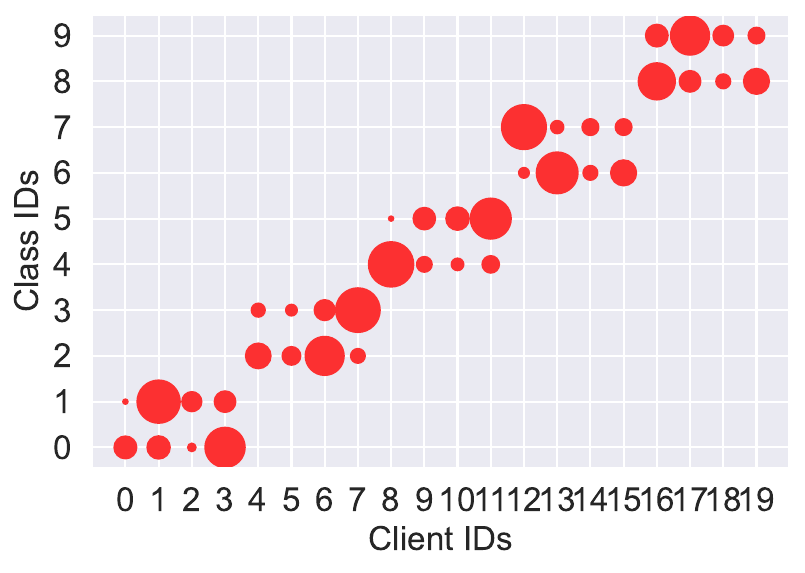}}
    \hfill
	\subfigure[Flowers102]{\includegraphics[width=0.48\linewidth]{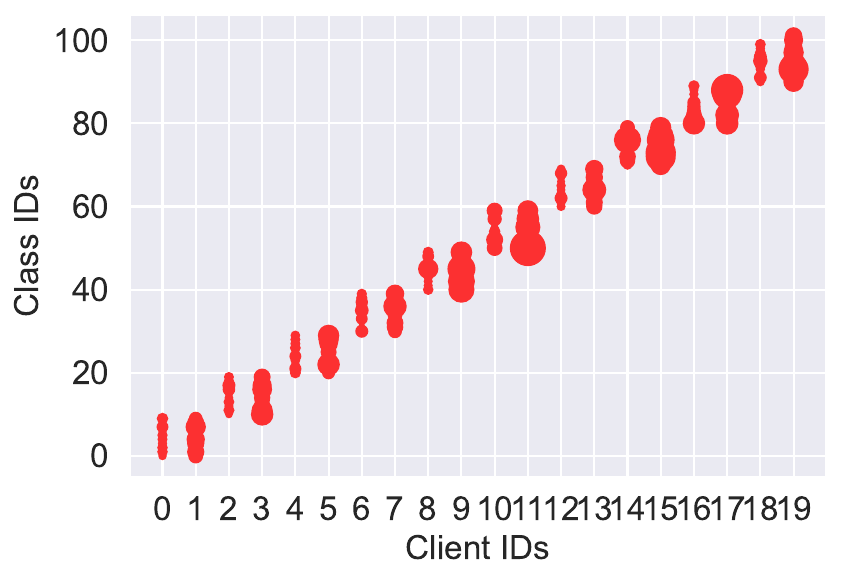}}
	\hfill
    \hfill
	\subfigure[Cifar100]{\includegraphics[width=0.48\linewidth]{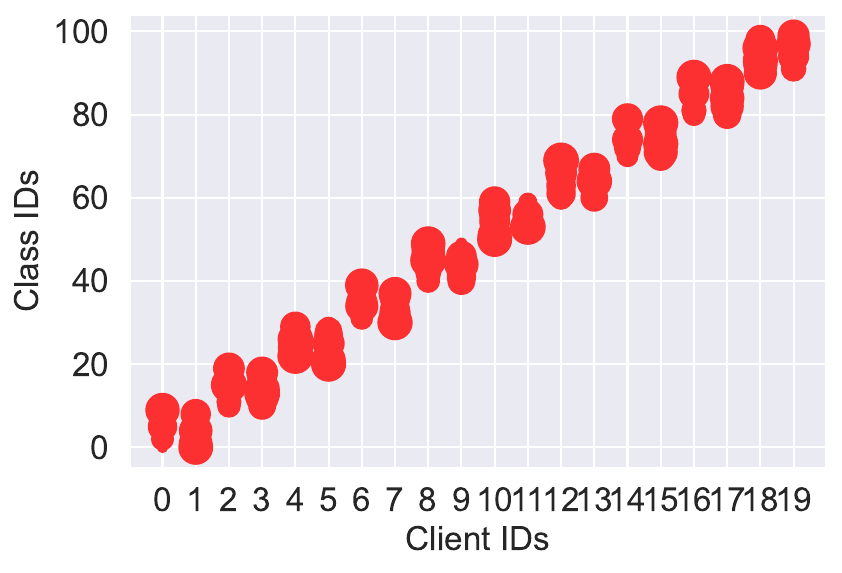}}
    \hfill
	\subfigure[Tiny-ImageNet]{\includegraphics[width=0.48\linewidth]{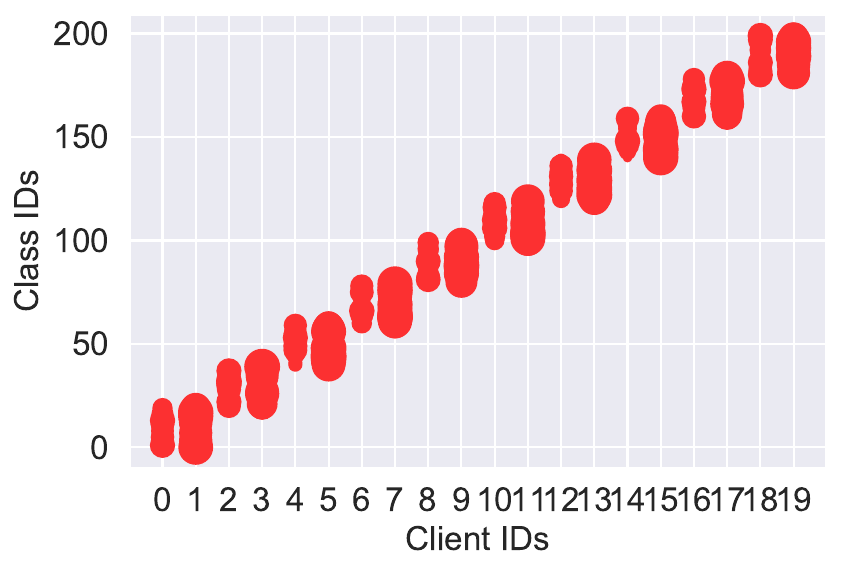}}
    \hfill
    \hfill
	\caption{The data distribution of each client on Cifar10, Flowers102, Cifar100, and Tiny-ImageNet, respectively, in the pathological settings. The size of a circle represents the number of samples. }
\end{figure*}

\begin{figure*}[ht]
	\centering
    \hfill
	\subfigure[Cifar10]{\includegraphics[width=0.48\linewidth]{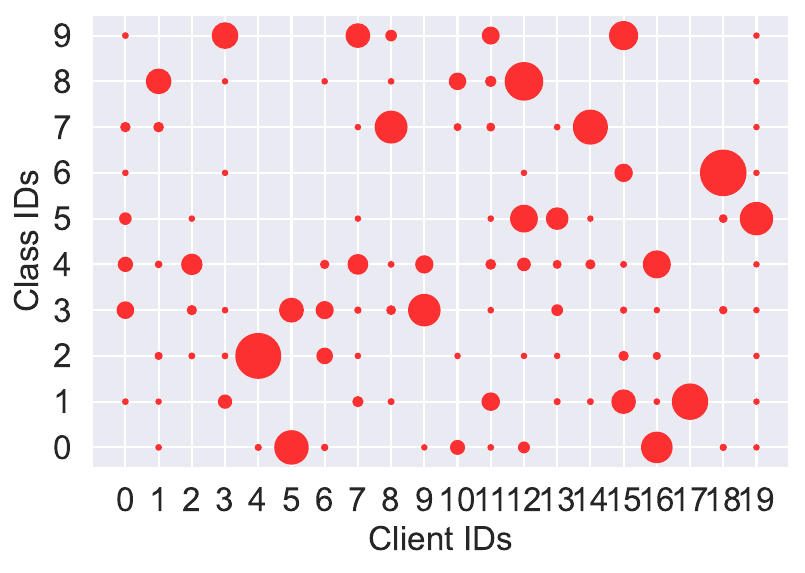}}
    \hfill
	\subfigure[Flowers102]{\includegraphics[width=0.48\linewidth]{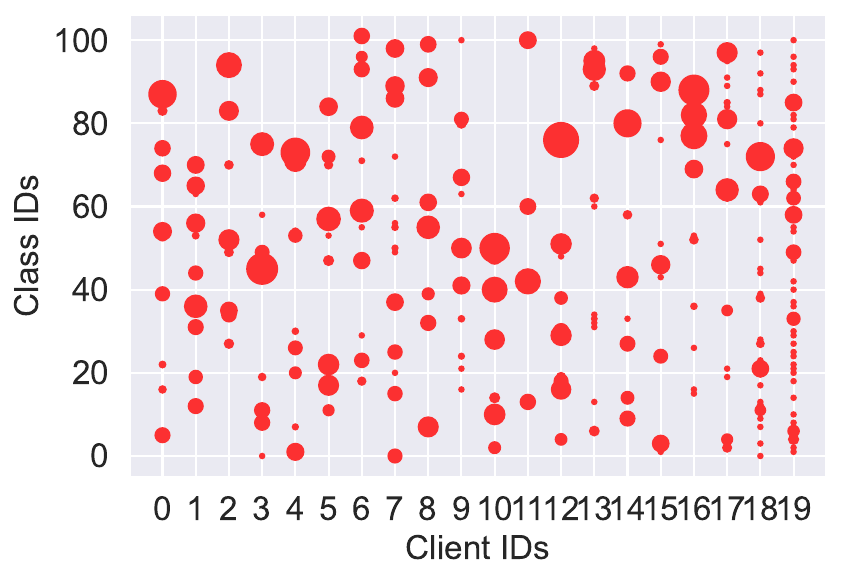}}
	\hfill
    \hfill
	\subfigure[Cifar100]{\includegraphics[width=0.48\linewidth]{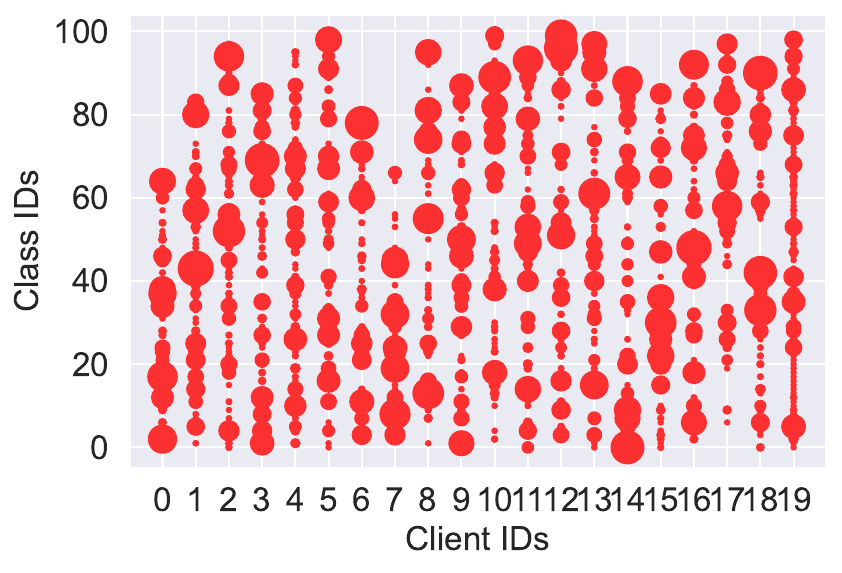}}
    \hfill
	\subfigure[Tiny-ImageNet]{\includegraphics[width=0.48\linewidth]{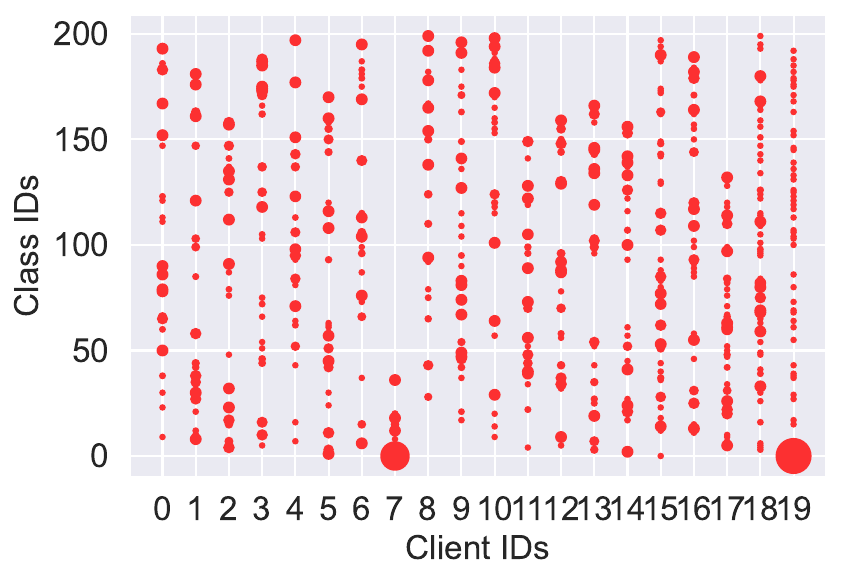}}
    \hfill
    \hfill
	\caption{The data distribution of each client on Cifar10 ($\beta=0.1$), Flowers102 ($\beta=0.01$), Cifar100 ($\beta=0.1$), and Tiny-ImageNet ($\beta=0.01$), respectively, in Dirichlet setting s. The size of a circle represents the number of samples. }
\end{figure*}

\begin{figure*}[ht]
	\centering
	\subfigure[50 clients]{\includegraphics[width=\linewidth]{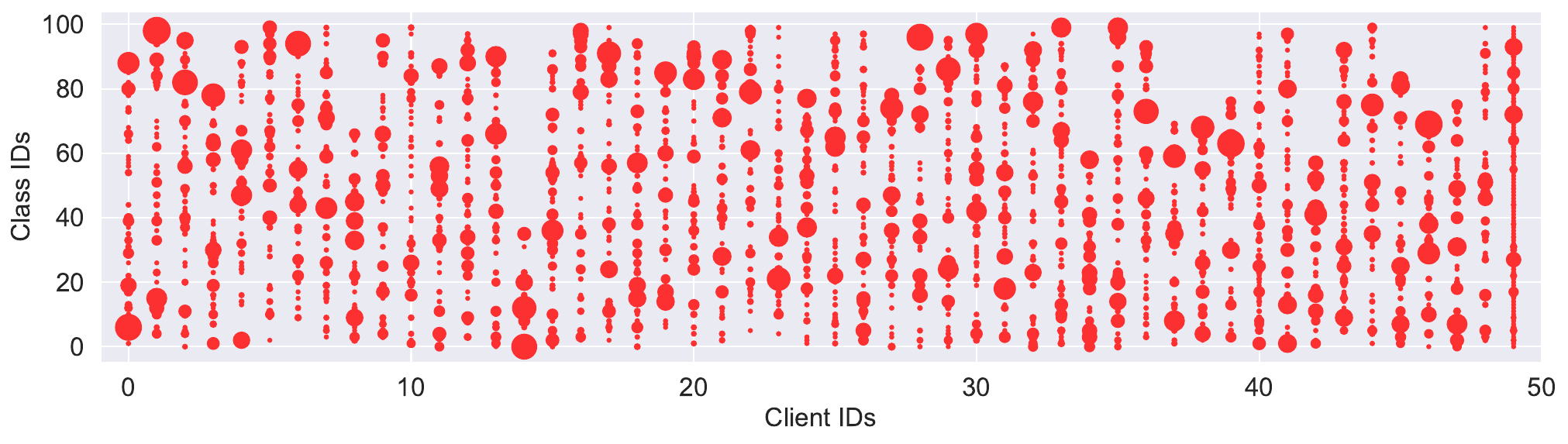}}
	\hfill
	\subfigure[100 clients]{\includegraphics[width=\linewidth]{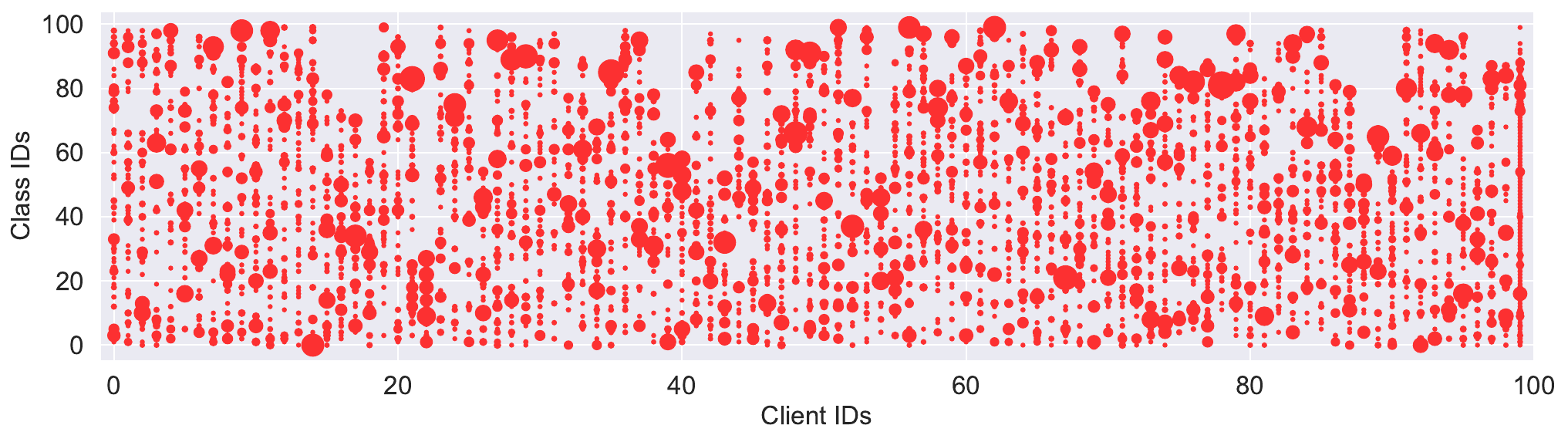}}
	\hfill
	\caption{The data distribution of each client on Cifar100 in the Dirichlet setting  ($\beta=0.1$) with 50 and 100 clients, respectively. The size of a circle represents the number of samples. }
\end{figure*}

\end{document}